\documentclass[10pt,twocolumn,letterpaper]{article}

\usepackage[pagenumbers]{cvpr} 

\definecolor{mustard}{RGB}{255, 219, 88} 

\usepackage{pifont}

\usepackage{makecell}
\usepackage{fontawesome5}
\usepackage[ruled,vlined]{algorithm2e}
\usepackage{xcolor} 
\newcommand{\Cmark}[1][green!60!black]{\textcolor{#1}{\ding{51}}} 
\newcommand{\Xmark}[1][red!70!black]{\textcolor{#1}{\ding{55}}}
\usepackage{booktabs}
\usepackage{multirow}  
\usepackage{booktabs}
\usepackage{array}
\usepackage{colortbl}

\definecolor{cvprblue}{rgb}{0.21,0.49,0.74}
\usepackage[breaklinks,colorlinks,allcolors=cvprblue]{hyperref}
\usepackage[accsupp]{axessibility}  
\usepackage{booktabs}
\usepackage{graphicx}
\usepackage[table]{xcolor}
\usepackage{xcolor}
\definecolor{teal}{RGB}{0,128,128}
\definecolor{crimson}{RGB}{220,20,60}
\definecolor{royalblue}{RGB}{65,105,225}
\definecolor{darkviolet}{RGB}{148,0,211}

\def\paperID{10732} 
\def\confName{CVPR}
\def\confYear{2026}

\title{Revisiting 2D Foundation Models for Scalable 3D Medical Image Classification}

\author{
Han Liu$^{1}$\thanks{Corresponding author: han.liu@siemens-healthineers.com}
\quad
Bogdan Georgescu$^{1}$
\quad
Yanbo Zhang$^{1}$
\quad
Youngjin Yoo$^{1}$
\quad
Michael Baumgartner$^{2}$
\\
Riqiang Gao$^{1}$
\quad
Jianing Wang$^{1}$
\quad
Gengyan Zhao$^{1}$
\quad
Eli Gibson$^{1}$
\quad
Dorin Comaniciu$^{1}$
\quad
Sasa Grbic$^{1}$
\\
\\
$^{1}$Digital Technology and Innovation, Siemens Healthineers, Princeton NJ, USA
\\
$^{2}$Digital Technology and Innovation, Siemens Healthineers, Erlangen, Germany
}

\begin{document}
\maketitle

\begin{abstract}

3D medical image classification is essential for modern clinical workflows. Medical foundation models (FMs) have emerged as a promising approach for scaling to new tasks, yet current research suffers from three critical pitfalls: data-regime bias, suboptimal adaptation, and insufficient task coverage. In this paper, we address these pitfalls and introduce \textbf{AnyMC3D}, a scalable 3D classifier adapted from 2D FMs. Our method scales efficiently to new tasks by adding only lightweight plugins (about 1M parameters per task) on top of a single frozen backbone. This versatile framework also supports multi-view inputs, auxiliary pixel-level supervision, and interpretable heatmap generation. We establish a comprehensive benchmark of 12 tasks covering diverse pathologies, anatomies, and modalities, and systematically analyze state-of-the-art 3D classification techniques. Our analysis reveals key insights: (1) effective adaptation is essential to unlock FM potential, (2) general-purpose FMs can match medical-specific FMs if properly adapted, and (3) 2D-based methods surpass 3D architectures for 3D classification. For the first time, we demonstrate the feasibility of achieving state-of-the-art performance across diverse applications using a single scalable framework (including 1st place in the VLM3D challenge), eliminating the need for separate task-specific models.

\end{abstract}    
\section{Introduction}
\label{sec:intro}

3D medical image classification plays an essential role in clinical workflows, such as emergency triage \cite{yoo2025non}, disease diagnosis \cite{alves2022fully} and severity grading \cite{chen2019fully}. As clinical needs continue to expand, \textbf{scalability}, i.e., rapid model deployment with minimal training effort, has become a key property of modern deep learning models. However, in the medical domain, scalability is often constrained by the limited training data and stringent clinical accuracy requirements. Given a new task, traditional methods extend 2D architectures (e.g., ResNet \cite{he2016deep} and DenseNet \cite{huang2017densely}) to 3D to capture inter-slice dependencies. However, 3D models are typically trained from scratch due to scarce pretrained weights. This causes severe overfitting on small per-task datasets, which is particularly problematic when scaling to new tasks where only limited initial data is available. More importantly, this one-model-per-task paradigm scales poorly as it requires separate model training for each new application.

\begin{figure}[t]
    \centering
    \includegraphics[width=1\linewidth]{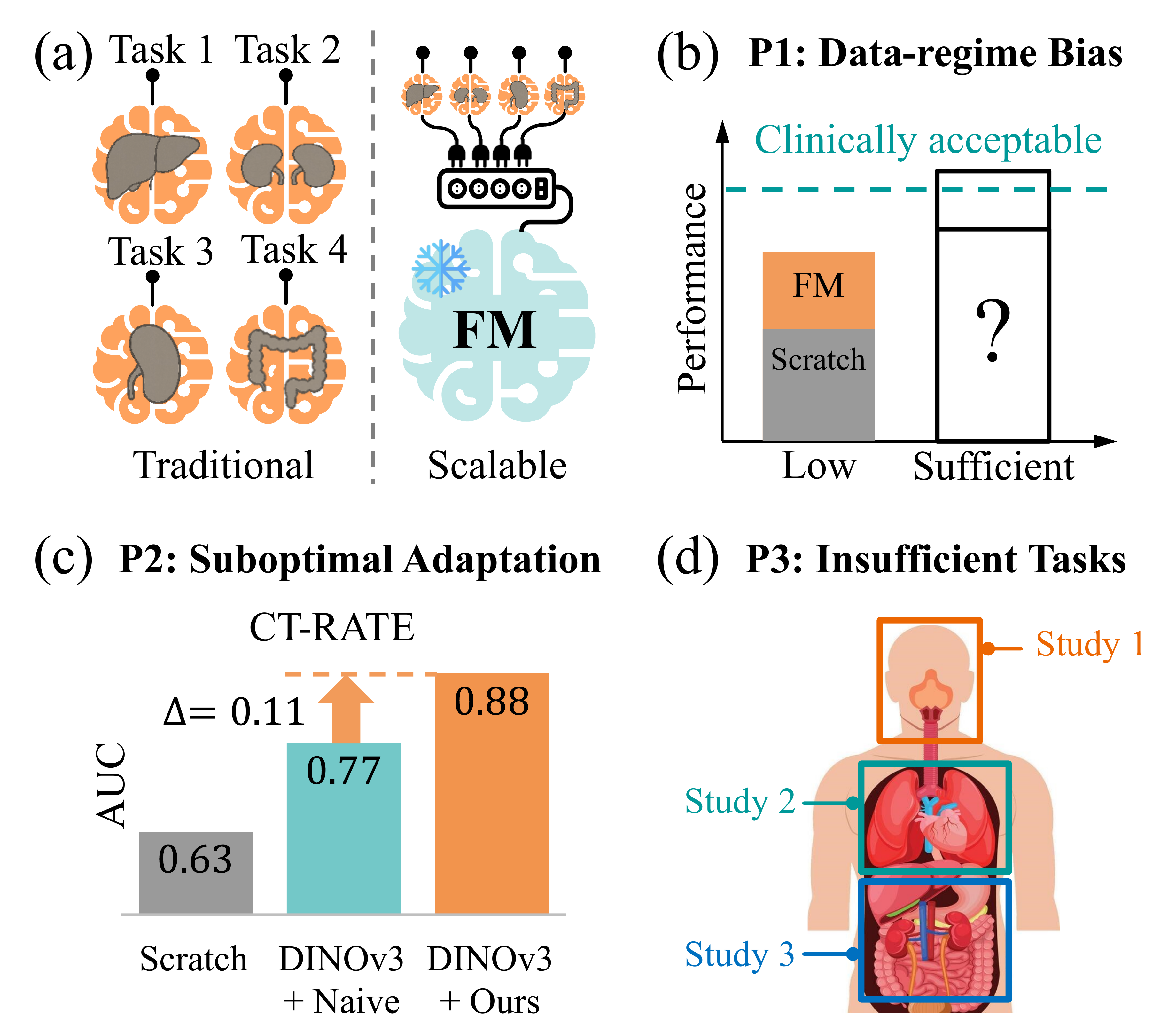}
            \caption{(a) Comparison of traditional and scalable methods. \textbf{Left}: traditional methods train task-specific models and thus scale poorly. \textbf{Right}: scalable methods only require adding lightweight task-specific plugins on a frozen FM. (b)-(d): Illustration of three common pitfalls in previous research. \textbf{P1}: Evaluation bias toward low-data regimes limits clinical relevance due to large accuracy gaps. \textbf{P2}: Suboptimal adaptation to downstream tasks limits the potential of FMs. \textbf{P3}: Insufficient task coverage across diverse pathologies, imaging modalities and anatomical regions.}
    \label{fig1}
\end{figure}

Recent advances in foundation models (FMs) offer a promising path for scaling to new tasks. Through large-scale pretraining, FMs learn robust visual features that can be efficiently adapted to diverse downstream applications. In recent years, many medical FMs have been developed by leveraging massive unlabeled medical data. 3D medical FMs~\cite{chen2019med3d,wu2024voco,hamamci2026generalist} are considered the standard approach for 3D tasks, as pretrained encoders naturally handle volumetric inputs. Recent work on 2D medical FMs~\cite{codella2024medimageinsight} demonstrates feasibility for 3D adaptation, while other studies~\cite{zhang2025benchmarking,liu2025does,baharoon2023evaluating} benchmark general-purpose 2D FMs such as DINO~\cite{oquab2023dinov2,simeoni2025dinov3} on 3D medical tasks. Despite this progress, we identify three critical pitfalls in existing research.

\textbf{P1 - Data-Regime Bias.} Prior work~\cite{ma2023foundation,zhu20253d,wang2023real,ji2025generative} predominantly evaluates medical FMs in low-data regimes such as few-shot fine-tuning. Adapted FMs typically show advantages over from-scratch baselines, but performance remains substantially below clinically acceptable levels (Fig.~\ref{fig1}b). This setup neglects evaluation with sufficient training data, failing to reveal performance at realistic dataset sizes. Some studies~\cite{napravnik2025lessons,taleb20203d,espis2025comparative} observe that FM benefits diminish as training data increases. Therefore, evaluation with adequate training data is essential to assess practical value for real-world deployment.

\textbf{P2 - Suboptimal Adaptation.} 
Prior FM studies primarily focus on model development and overlook the importance of proper adaptation to downstream tasks. For instance, the most common strategy of adapting 2D FMs to 3D is to use frozen 2D backbones as feature extractors and combine slice embeddings with average \cite{zhang2025benchmarking,liu2025does,baharoon2023evaluating} or median pooling \cite{codella2024medimageinsight}. However, we find that substantial performance gains ($\Delta$AUC=0.11) can be achieved with the same DINOv3 backbone by simply replacing this strategy \cite{liu2025does} with our method (Fig.~\ref{fig1}c). This indicates that FMs may have been underestimated in previous studies, and they must be properly adapted to unleash their potential.

\textbf{P3 - Insufficient Tasks.} 
Existing studies evaluate FMs on limited 3D classification tasks, typically within a single imaging modality or anatomical region. This narrow scope prevents comprehensive assessment of generalization across different modalities (CT, MRI), anatomical regions (head, chest, abdomen), and clinical applications. Without a benchmark of diverse 3D classification tasks, it remains unclear whether FMs provide robust, scalable solutions or only perform well on certain tasks.


In this paper, we address these pitfalls by investigating effective FM adaptation strategies and establishing a diverse benchmark spanning real-world 3D clinical applications. Our main contributions are as follows:

\begin{enumerate}
  \item \textbf{Identification of Critical Pitfalls.} We identify three critical pitfalls in prior studies: data-regime bias, suboptimal adaptation, and insufficient tasks. These pitfalls obscure accurate assessment of the potential of existing FMs and their values for real-world deployment.
  
  \item \textbf{Simple Yet Effective Method.} To address P2, we propose AnyMC3D, a scalable 3D classifier adapted from 2D FMs for \underline{\textbf{any}} \underline{\textbf{3D}} \underline{\textbf{m}}edical \underline{\textbf{c}}lassification task. For the first time, we demonstrate the feasibility of achieving superior performance across diverse applications using a single scalable framework, eliminating the need for separate task-specific 3D models.

  \item \textbf{Comprehensive Benchmark and Extensive Evaluation.} To address P1 and P3, we establish a benchmark of 12 diverse tasks with realistic dataset sizes, spanning multiple imaging modalities and anatomical regions. We also systematically evaluate the existing 3D classification approaches with in-depth analyses.
  
  \item \textbf{Key Findings and Critical Insights.} Through extensive analysis, we reveal findings that challenge widely-accepted practices in medical FM research, providing actionable insights for future work.
\end{enumerate}
\section{Related Work}
\label{sec:related}
\subsection{Medical Foundation Models}
\label{sec:mfm}
Recent medical FMs have emerged as a promising solution to longstanding issues of data scarcity and poor generalization. Existing studies have mainly focused on \textit{building better pretrained models}~\cite{moor2023foundation,wu2025towards,wu2024voco,chen2019med3d,codella2024medimageinsight,sellergren2025medgemma,amadou2024echoapex,ma2025fully,hamamci2026generalist}. These models typically differ in two aspects: (1) \textbf{\textit{Pretraining objectives}}: some works apply the existing pretraining approaches in computer vision to medical datasets, such as supervised pretraining~\cite{ma2025fully,li2024well,ma2023foundation}, self-supervised pretraining~\cite{chen2024towards,amadou2024echoapex,chen2019med3d,ghesu2022contrastive,vorontsov2023virchow} and vision-language pretraining~\cite{codella2024medimageinsight,sellergren2025medgemma,hamamci2026generalist,blankemeier2026merlin}. Additionally, some studies design novel pretraining objectives that are more tailored to medical imaging~\cite{zhou2021models,tang2022self,wu2024voco,xu2025generalizable}. (2) \textbf{\textit{Scope}}: the scope of FMs may vary from body regions such as brain~\cite{yoo2025non,zhu20253d} and chest~\cite{hamamci2026generalist,ji2025generative,niu2025medical}, to imaging modalities such as CT~\cite{blankemeier2026merlin,zhu20253d} and MRI~\cite{dong2025mri,tak2024foundation}, to comprehensive medical images~\cite{codella2024medimageinsight,sellergren2025medgemma}. In contrast to prior work on developing new FMs, our study focuses on optimizing existing FMs through effective adaptation.

\subsection{3D Medical Image Classification}
3D medical image classification is challenging due to volumetric data complexity and limited labeled datasets. Existing approaches fall into two categories with distinct trade-offs:
(1) \textbf{\textit{Performance}}: early deep learning methods often extend 2D architectures to 3D and train from-scratch, but they only achieve good performance when massive training data are available. More recent methods, such as M3T~\cite{jang2022m3t} and MST~\cite{muller2025medical}, leverage the pretrained 2D models on natural images through transfer learning combined with slice fusion. Similarly, winning solutions~\cite{rsna2022kaggle,rsna2023kaggle} in recent 3D classification challenges~\cite{lin2023rsna,rudie2024rsna} also adopt this strategy. Despite the superior performance, these methods suffer from limited scalability due to full fine-tuning requirements. (2) \textbf{\textit{Scalability}}: In recent years, medical foundation models (FMs) have emerged as scalable solutions for rapid model development. Several studies~\cite{ghesu2022contrastive,chen2019med3d,wu2024voco} demonstrate that 3D FMs can be efficiently specialized to new tasks through lightweight adaptation. By contrast, 2D FMs have been primarily adopted for 2D tasks, with their potential for 3D classification underexplored. Existing studies either employ suboptimal adaptation strategies~\cite{liu2025does,zhang2025benchmarking,codella2024medimageinsight} or focus exclusively on 2D evaluation~\cite{sellergren2025medgemma}. Liu et al.~\cite{liu2025does} evaluate DINOv3 on CT-RATE~\cite{hamamci2026generalist} by extracting slice embeddings with a frozen backbone followed by average pooling. Zhang et al.~\cite{zhang2025benchmarking} adopt the same strategy for stroke classification, while Codella et al.~\cite{codella2024medimageinsight} use frozen MedImageInsight with median pooling for 3D retrieval. While using frozen 2D FMs is highly scalable, this approach significantly limits performance as generic pretrained features struggle to capture subtle task-specific medical findings. More recently, Veasey et al.~\cite{veasey2025low} apply LoRA~\cite{hu2022lora}-adapted DINOv2 for lung nodule classification. However, their approach treats 3D volumes as 2D by using only three orthogonal slices as input, failing to leverage the full 3D spatial context.

\section{Methodology}
\label{sec:method}

\subsection{Problem Formulation}
\label{prelim}
In 3D medical image classification, an input image is denoted as a 4D tensor $\mathbf{x}\in\mathbb{R}^{C\times H\times W\times S}$, where $C$ is the number of channels and $(H,W,S)$ are the spatial dimensions. Given a dataset $\mathcal{D}=\{(\mathbf{x}_i,y_i)\}_{i=1}^N$, $y_i\in\{1,\ldots,K\}$ for multi-class classification, or $y_i\in\{0,1\}^K$ for multi-label classification, where $K$ denotes the number of target classes. A 3D classifier is a parametric mapping:
\begin{align}
f_\theta: \mathbb{R}^{C\times H\times W\times S}\ \longrightarrow\ \mathbb{R}^{K}
\end{align}
which returns a vector of class logits $\mathbf{z}=f_\theta(\mathbf{x})\in\mathbb{R}^{K}$. The parameters $\theta$ are learned via empirical risk minimization with a suitable classification loss $\ell$ applied to logits:
\begin{align}
\theta^\star \;=\; \arg\min_{\theta}\; \frac{1}{N}\sum_{i=1}^N \ell\!\big(f_\theta(\mathbf{x}_i), y_i\big)
\end{align}

\begin{figure}[t]
    \centering
    \includegraphics[width=1\linewidth]{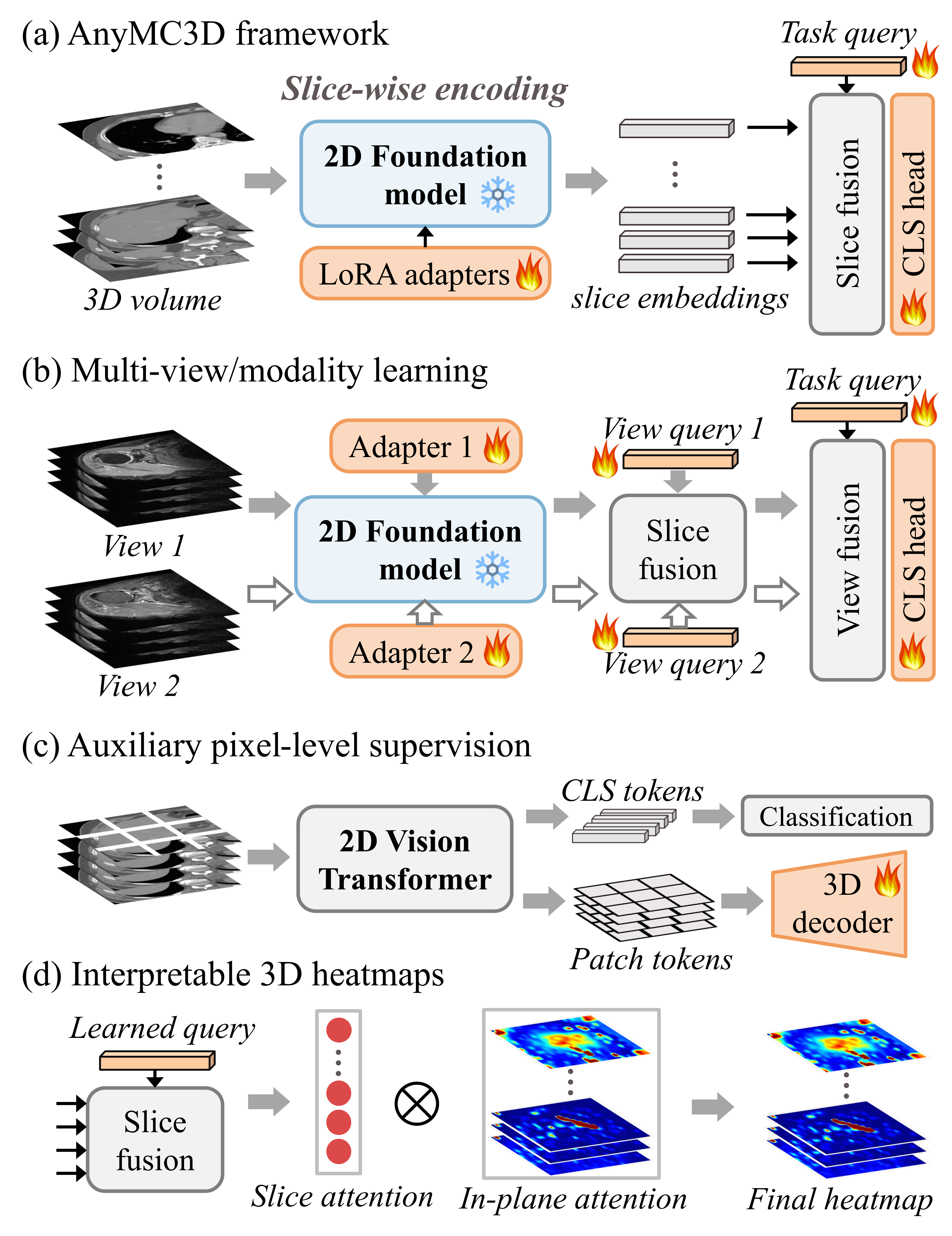}
            \caption{(a) Overall framework of AnyMC3D. To scale to a new task, only task-specific plugins (orange) need to be added while the 2D FM remains frozen. (b) Adaptive to arbitrary number of input views or sequences. (c) Flexible to train with auxiliary pixel-level supervision. (d) Generate interpretable 3D heatmaps.}
    \label{fig:method}
\end{figure}

\subsection{Proposed Method: AnyMC3D}
Fig. \ref{fig:method}a presents the overall framework of AnyMC3D, a scalable 3D classifier adapted from 2D FMs. It can efficiently scale to new tasks by adding only lightweight, task-specific plugins on top of a frozen 2D FM backbone.

\paragraph{In-Plane Reasoning with Adapted 2D FM.} To leverage 2D FMs for 3D data, we decouple in-plane feature extraction from through-plane reasoning. To scale to a new task $t$, we freeze the entire 2D backbone $f_\theta$ and adapt it with LoRA adapters $\psi_t$. Specifically, we apply LoRA to the patch embedding and all self-attention projection layers. We freeze the pretrained weight matrix $\mathbf{W}\!\in\!\mathbb{R}^{d_{\mathrm{in}}\times d_{\mathrm{out}}}$ and learn a task-specific, low-rank update $\Delta\mathbf{W_t}$:
\begin{align}
\mathbf{W}' = \mathbf{W} + \Delta\mathbf{W_t}, 
\quad 
\Delta\mathbf{W_t} = \tfrac{\alpha}{r}\,\mathbf{B_t}\mathbf{A_t}
\end{align}
where $\mathbf{B_t}\in\mathbb{R}^{d_{\mathrm{in}}\times r}$ and $\mathbf{A_t}\in\mathbb{R}^{r\times d_{\mathrm{out}}}$ are the low-rank matrices. Rank $r\ll \min(d_{\mathrm{in}},d_{\mathrm{out}})$ controls learnable capacity and $\alpha$ scales update magnitude. Following~\cite{hu2022lora}, we use zero initialization for $\mathbf{B_t}$ to preserve the pretrained behaviors of FMs at the beginning of training.

Let the 3D input be \(\mathbf{x}\in\mathbb{R}^{C\times H\times W\times S}\), with \(S\) slices along a chosen axis. We form 2D slices \(\mathbf{x}^{s}\in\mathbb{R}^{C\times H\times W}\), encode each slice with the adapted 2D FM, and extract the \textit{class}-token from the last block as the slice embedding:
\begin{align}\label{slice_embed}
\mathbf{h}_s=\tilde{f}_{\theta,\psi_t}\!\big(\mathbf{x}^{s}\big)\in\mathbb{R}^{d}, \quad s=1,\ldots,S
\end{align}

\paragraph{Permutation-Invariant Slice Aggregation.} 
To fuse slice embeddings, we design a lightweight module that maintains strong performance while ensuring scalability. Prior work typically imposes ordered priors through sequence modeling with RNNs~\cite{rsna2022kaggle,rsna2023kaggle} or Transformers~\cite{muller2025medical}. However, 3D medical images often exhibit anisotropic spacing and variable coverage. Strict sequence modeling can be overly prescriptive and sensitive to acquisition variability. Therefore, we propose to fuse slice embeddings in a permutation-invariant manner via query-based attention pooling.

Specifically, slice embeddings are stacked as \(\mathbf{H}=[\mathbf{h}_1,\ldots,\mathbf{h}_S]^\top\in\mathbb{R}^{S\times d}\) and aggregated using a learnable task query \(\mathbf{q}_t\in\mathbb{R}^{d}\) that assigns higher weights to task-relevant slices. The volume embedding $\mathbf{v}$ is computed as:
\begin{align}\label{attn_eq}
\boldsymbol{a} = \operatorname{softmax}\!\Big(\tfrac{\mathbf{H}\mathbf{q}_t}{\sqrt{d}}\Big)\in\mathbb{R}^{S}, 
\quad
\mathbf{v} = \boldsymbol{a}^{\top}\mathbf{H}\in\mathbb{R}^{d}
\end{align}
where \(d\) is the embedding dimension and \(\boldsymbol{a}\) are normalized attention weights. Finally, a classification head produces class logits: $\mathbf{z} = g_{\omega}(\mathbf{v})\in\mathbb{R}^{K}$.

\subsection{Versatility for 3D Medical Imaging}
To ensure broad applicability across various 3D medical imaging tasks, we design three essential extensions: (1) multi-view/modal input support, (2) auxiliary pixel-level supervision, and (3) interpretable heatmap generation. For clarity, task index $t$ is omitted unless necessary.

\paragraph{Multi-View Learning.} In many medical imaging studies, particularly MRI, each subject may include multiple views (sagittal, coronal) or sequences (T1, T2, FLAIR). AnyMC3D can be extended to harness such complementary cues through an efficient late fusion strategy. As shown in Fig.~\ref{fig:method}b, we encode each view separately with view-specific LoRA adapters and queries, then aggregate the resulting view embeddings for patient-level prediction. Formally, for view $i$, the view embedding $\mathbf{v}^{(i)}$ is computed using view-specific adapters $\psi^{(i)}$ and query $\mathbf{q}^{(i)}$ via Eq.~\ref{attn_eq}. The view embeddings are stacked and fused with task query $\mathbf{q}_t$ by attention pooling. Since only lightweight view-specific components are added per view (only the orange components are trainable in Fig.\ref{fig:method}b), this approach scales efficiently to arbitrary numbers of input views.

\paragraph{Auxiliary Pixel-Level Supervision.}
\label{aux} Training 3D classifiers with only image-level labels can be challenging for subtle findings and tiny objects, such as hemorrhage and pulmonary nodules. Pixel-level supervision provides precise spatial signals that disambiguate which regions support the class decision. Previous studies \cite{dzieniszewska2025deep,wang2021joint} show that additional pixel-level supervision (even if only available for a subset) can significantly improve classification accuracy. To support pixel-level supervision, AnyMC3D can be trained with a multi-task objective by imposing additional spatial regularization on the \textit{patch}-tokens.

As shown in Fig. \ref{fig:method}c, we extract the patch tokens from the last block \(\mathbf{P}_s\in\mathbb{R}^{N\times d}\) for slice $s$, where \(N=\tfrac{H}{P}\times \tfrac{W}{P}\) for patch size \(P\). We then reshape the patch tokens to a 2D feature map and stack all feature maps along the slice axis to form a pseudo\mbox{-}3D token volume:
\begin{align}
\mathbf{F}_s &\;=\; \operatorname{Reshape}(\mathbf{P}_s)\in\mathbb{R}^{d\times \tfrac{H}{P}\times \tfrac{W}{P}}
\end{align}
\begin{align}
\label{p3d}
\mathbf{F} &\;=\; \big[\mathbf{F}_1,\ldots,\mathbf{F}_S\big] \in \mathbb{R}^{d\times \tfrac{H}{P}\times \tfrac{W}{P}\times S}
\end{align}

Afterwards, we use a lightweight 3D decoder to map the pseudo 3D token volume back to voxel-wise logits \(\hat{\mathbf{Y}}\in\mathbb{R}^{K_{\mathrm{seg}}\times H\times W\times S}\). Let \(\mathcal{I}\) be the subset of training cases with segmentation masks \(\mathbf{Y}\). The training objective couples the classification loss \(\mathcal{L}_{\mathrm{cls}}\) with an auxiliary segmentation loss applied only on \(\mathcal{I}\):
\begin{align}
\mathcal{L}_{\mathrm{total}}
\;=\;
\mathcal{L}_{\mathrm{cls}}
\;+\;
\lambda_{\mathrm{seg}} \cdot \frac{1}{|\mathcal{I}|}\sum_{i\in\mathcal{I}}\mathcal{L}_{\mathrm{seg}}\!\big(\hat{\mathbf{Y}}_i,\mathbf{Y}_i\big)
\end{align}
where \(\lambda_{\mathrm{seg}}>0\) balances the auxiliary signal. At inference time, the segmentation branch can be omitted to avoid extra computation cost for classification.

\paragraph{Interpretable Heatmaps.} Interpretable saliency maps can help verify plausible anatomical attention, reveal failure modes, and support reader trust. Previous studies \cite{abnar2020quantifying,muller2025medical} demonstrate that vision transformers can generate explainable 2D heatmaps by visualizing the attention between the class token and patch tokens. Inspired by this, we propose to generate 3D heatmaps by combining per-slice 2D heatmaps with the corresponding slice importance scores.

\begin{table*}[t]
\begin{minipage}[t]{0.32\textwidth}
    \centering
    \vspace{0pt}
    \includegraphics[width=\textwidth]{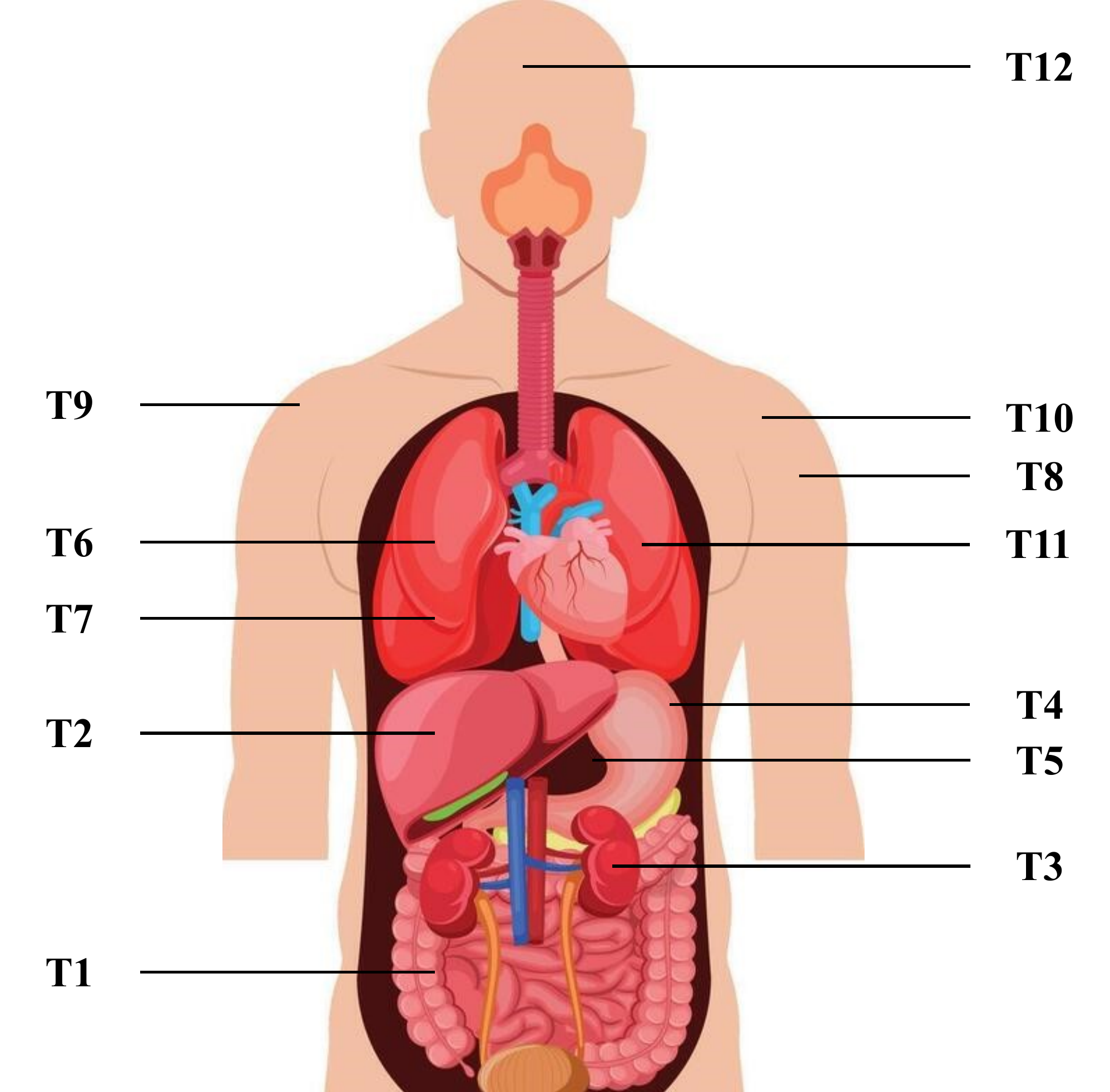}
    \label{dataset}
\end{minipage}
\hfill
\begin{minipage}[t]{0.62\textwidth}
    \vspace{0pt}
    \captionof{table}{Summary of the datasets used for evaluation across 12 tasks (T1-T12).}
    \label{dataset_summary}
    \footnotesize
    \setlength{\tabcolsep}{4pt}  
    \begin{tabular}{lllcllc}
    \toprule
    Task & Label & Modality & Views & Dataset & Region & Volumes \\
    \midrule\midrule
    \rowcolor{white} T1 & Bowel Injury & CECT & 1 & RATIC \cite{rudie2024rsna} & Abdomen & 4,679  \\
    \rowcolor{gray!15} T2 & Liver Injury & CECT & 1 & RATIC \cite{rudie2024rsna} & Abdomen & 4,701 \\
    \rowcolor{white} T3 & Kidney Injury & CECT & 1 & RATIC \cite{rudie2024rsna} & Abdomen & 4,677 \\
    \rowcolor{gray!15} T4 & Spleen Injury & CECT & 1 & RATIC \cite{rudie2024rsna} & Abdomen & 4,695 \\
    \rowcolor{white} T5 & PDAC & CECT & 1 & PANORAMA \cite{alves2024panorama} & Abdomen & 2,238 \\
    \rowcolor{gray!15} T6 & Nodule Malig. & CT & 1 & Private & Chest & 1,140 \\
    \rowcolor{white} T7 & Nodule Spicu. & CT & 1 & Private & Chest & 5,668 \\
    \rowcolor{gray!15} T8 & Bicep Tear & MRI & 2 & Private & Shoulder & 12,159 \\
    \rowcolor{white} T9 & Bursa Fluid & MRI & 3 & Private & Shoulder & 10,978 \\
    \rowcolor{gray!15} T10 & Labrum Tear & MRI & 2 & Private & Shoulder & 12,191 \\
    \rowcolor{white} T11 & Chest Multi-Abn. & CT & 1 & CT-RATE \cite{hamamci2026generalist} & Chest & 50,188 \\
    \rowcolor{gray!15} T12 & Head Multi-Abn. & CT & 1 & Private & Head & 29,476 \\
    \bottomrule
    \end{tabular}
\end{minipage}
\end{table*}

As shown in Fig. \ref{fig:method}d, we first extract the in-plane attention maps for each slice. Consider the last transformer block \(L\) with \(h\) heads and head dimension \(d_h\). For slice \(s\), queries and keys of head \(j\) are \(\mathbf{Q}^{(L,j)}_s,\mathbf{K}^{(L,j)}_s\in\mathbb{R}^{(1+R+N)\times d_h}\), where the sequence contains one class token, \(R\) optional register tokens, and \(N\) patch tokens. For each head, the class\mbox{-}to\mbox{-}patch attention is computed as:
\begin{align}
\mathbf{A}^{(L,j)}_s \;=\; \operatorname{softmax}\!\Big(\tfrac{\mathbf{Q}^{(L,j)}_s\,\mathbf{K}^{(L,j)\top}_s}{\sqrt{d_h}}\Big)
\end{align}
The attention is then averaged over heads, reshaped to the patch grid, and upsampled to the original image dimension.
\begin{align}
\mathbf{m}_s &\;=\;\tfrac{1}{h}\sum_{j=1}^{h}\big(\mathbf{A}^{(L,j)}_s\big)_{\mathrm{cls},\,\mathrm{patch}}\in\mathbb{R}^{N}
\end{align}
\begin{align}
\mathcal{M}_s &= \operatorname{Up}(\operatorname{Reshape}\!\big(\mathbf{m}_s\big)) \in\mathbb{R}^{H\times W}
\end{align}

Then we compute the importance score of each slice with the learned task query, as in Eq.~\ref{attn_eq}. Lastly, the 3D heatmap is obtained by stacking the 2D heatmaps weighted by the corresponding slice importance scores.

\section{Experiments}
\label{sec:experiments}
In this section, we first present the implementation details and our benchmark. Then, we describe the state-of-the-art baselines for 3D medical classification. Finally, we report detailed experimental results with in-depth analysis.

\subsection{Experiment Setup}
\paragraph{Implementation Details.} We implement AnyMC3D with three existing 2D FMs, including two medical FMs, MedImageInsight (MII)~\cite{codella2024medimageinsight} and MedGemma~\cite{sellergren2025medgemma}, and a general-purpose FM, DINO~\cite{oquab2023dinov2,simeoni2025dinov3}. The vision encoders of MII and MedGemma are DaViT~\cite{ding2022davit} (365M) and SigLIP~\cite{zhai2023sigmoid} (432M), respectively. To keep backbone size comparable, we use the pretrained ViT-L (300M) from DINOv2~\cite{oquab2023dinov2} and DINOv3~\cite{simeoni2025dinov3}. We empirically set the LoRA rank and scaling factor as 8 and 16, respectively. For quantitative evaluation, we use the Area Under the Receiver Operating Characteristic curve (AUROC). More implementation details are provided in Appendix~\ref{app.a}.

\paragraph{Datasets and Tasks.}
To address P3, we establish a benchmark of 12 diverse tasks across body regions, pathologies, and modalities (Tab.~\ref{dataset_summary}). To address P1, we use realistic dataset sizes preserving natural class imbalance, where patients with disease are often outnumbered by healthy controls. For example, bowel injury has only 104 positive samples from 4,679 volumes. See Appendix~\ref{app.b} for details.

\paragraph{Competing Methods.} We compare AnyMC3D against the state-of-the-art 3D classification methods, including: \textbf{(1) From-scratch 3D backbones}: 3D ResNet~\cite{he2016deep}, 3D DenseNet~\cite{huang2017densely} and 3D ConvNeXt~\cite{liu2022convnet}. \textbf{(2) 2D/2.5D backbones with slice fusion}: M3T~\cite{jang2022m3t}: a transformer-based classifier that fuses the features encoded from 3 orthogonal directions. MST~\cite{muller2025medical}: a slice transformer that fully finetunes the DINOv2 encoder. RSNA-Kaggle: the winning solution of the RSNA-Kaggle 3D classification challenges~\cite{rsna2022kaggle,rsna2023kaggle,rsna2024kaggle}, which finetunes 2D pretrained backbones with 2.5D input and uses bidirectional-LSTM for slice fusion. We follow~\cite{rsna2023kaggle} to use EfficientNet as backbone and remove their segmentation branch for fair comparison. \textbf{(3) 2D medical FMs}: MII~\cite{codella2024medimageinsight} and MedGemma~\cite{sellergren2025medgemma} are 2D medical FMs pretrained with diverse, large-scale medical images. We freeze their backbones to extract slice embeddings and apply our slice fusion method. \textbf{(4) 3D medical FMs}: MedicalNet~\cite{chen2019med3d}: a 3D ResNet50 pretrained on large-scale medical images. VoCo~\cite{wu2024voco}: a Swin-UNETR model pretrained on 3D medical images. We report their full finetuning results as they substantially outperform LoRA adaptation and linear probing in preliminary experiments (Appendix~\ref{app.c}). \textbf{(5) Task-optimized baselines}: competitive approaches specifically optimized for the evaluated tasks. Detailed descriptions of baselines are in Appendix~\ref{app.c}.

\begin{table*}[t]
\centering
\caption{Comparison against the state-of-the-art 3D classification methods across 10 tasks. Best results are shown in \textbf{bold}, second best are \underline{underlined}. Train.Param.: trainable parameters (M). Fro.: frozen backbone. \textcolor{teal}{$\bullet$} 3D backbones; \textcolor{crimson}{$\bullet$} 2D/2.5D backbones (pretrained on natural images) + slice fusion; \textcolor{darkviolet}{$\bullet$} 3D medical FMs; \textcolor{royalblue}{$\bullet$} 2D medical FMs + slice fusion; \textcolor{mustard}{$\bullet$} AnyMC3D.}
\label{tab:classification_auc}
\resizebox{\textwidth}{!}{
\setlength{\arrayrulewidth}{0.4pt}%
\arrayrulecolor{black}%
\begin{tabular}{l!{\vrule}c!{\vrule}c!{\vrule}*{10}{>{\centering\arraybackslash}m{0.05\textwidth}}!{\vrule}*{2}{>{\centering\arraybackslash}m{0.05\textwidth}}}
\toprule
Method & \makecell{Train.\\Param.} & Fro. & Bowel (T1) & Liver (T2) & Kidney (T3) & Spleen (T4) & PDAC (T5) & Malig. (T6) & Spic. (T7) & Bicep (T8) & Bursa (T9) & Labr. (T10) & Avg. AUC & Avg. Rank \\
\midrule\midrule
\rowcolor{white} \textcolor{teal}{$\bullet$} 3D ResNet \cite{he2016deep}& 33.14 & \Xmark & 0.837 & 0.756 & 0.904 & 0.922 & 0.892 & 0.673 & 0.897 & 0.699 & 0.826 & 0.686 & 0.809 & 7.6 \\
\rowcolor{gray!15} \textcolor{teal}{$\bullet$} 3D DenseNet \cite{huang2017densely}& 11.24 & \Xmark & 0.926 & 0.842 & 0.951 & 0.925 & 0.916 & 0.654 & 0.903 & 0.724 & 0.849 & 0.637 & 0.833 & 6.4 \\
\rowcolor{white} \textcolor{teal}{$\bullet$} 3D ConvNeXt \cite{liu2022convnet}& 31.32 & \Xmark & 0.641 & 0.729 & 0.604 & 0.817 & 0.709 & 0.567 & 0.857 & 0.622 & 0.698 & 0.587 & 0.683 & 11.5 \\
\rowcolor{gray!15} \textcolor{crimson}{$\bullet$} M3T \cite{jang2022m3t}& 29.23 & \Xmark & 0.790 & 0.827 & 0.831 & 0.923 & 0.899 & 0.673 & \textbf{0.908} & 0.702 & 0.838 & 0.658 & 0.805 & 7.5 \\
\rowcolor{white} \textcolor{crimson}{$\bullet$} MST \cite{muller2025medical}& 23.05 & \Xmark & \underline{0.956} & 0.907 & 0.975 & 0.933 & 0.920 & 0.677 & 0.834 & 0.852 & 0.859 & 0.778 & \underline{0.869} & 4.0 \\
\rowcolor{gray!15} \textcolor{crimson}{$\bullet$} RSNA-Kaggle \cite{rsna2023kaggle}& 25.04 & \Xmark & 0.659 & 0.836 & 0.980 & \textbf{0.957} & \underline{0.957} & 0.601 & 0.864 & 0.782 & 0.882 & 0.584 & 0.810 & 6.8 \\
\rowcolor{white} \textcolor{darkviolet}{$\bullet$} MedicalNet \cite{chen2019med3d} & 46.16 & \Xmark & 0.781 & 0.744 & 0.873 & 0.899 & 0.861 & 0.624 & 0.887 & 0.679 & 0.801 & 0.676 & 0.783 & 8.1 \\
\rowcolor{gray!15} \textcolor{darkviolet}{$\bullet$} VoCo \cite{wu2024voco}& 50.49 & \Xmark & 0.664 & 0.688 & 0.939 & 0.919 & 0.926 & 0.590 & 0.899 & 0.746 & 0.852 & 0.702 & 0.793 & 7.0 \\
\rowcolor{white} \textcolor{royalblue}{$\bullet$} MII \cite{codella2024medimageinsight}& 0.03 & \Cmark & 0.861 & 0.797 & 0.881 & 0.831 & 0.891 & 0.614 & 0.831 & 0.715 & 0.701 & 0.727 & 0.785 & 8.7 \\
\rowcolor{gray!15} \textcolor{royalblue}{$\bullet$} MedGemma \cite{sellergren2025medgemma}& 0.03 & \Cmark & 0.697 & 0.662 & 0.729 & 0.753 & 0.740 & 0.595 & 0.754 & 0.669 & 0.652 & 0.649 & 0.690 & 10.9 \\
\midrule
\rowcolor{white} \textcolor{mustard}{$\bullet$} AnyMC3D (MII) & 1.32 & \Cmark & \textbf{0.985} & \textbf{0.939} & \textbf{0.988} & \textbf{0.957} & \underline{0.957} & \underline{0.678} & 0.888 & \textbf{0.865} & \underline{0.889} & \textbf{0.795} & \textbf{0.894} & \textbf{1.7} \\
\rowcolor{gray!15} \textcolor{mustard}{$\bullet$} AnyMC3D (MedGemma) & 2.01 & \Cmark & 0.951 & 0.865 & 0.967 & 0.942 & 0.934 & 0.595 & 0.897 & 0.840 & 0.874 & \textbf{0.795} & 0.866 & 5.2 \\
\rowcolor{white} \textcolor{mustard}{$\bullet$} AnyMC3D (DINOv3) & 1.20 & \Cmark & 0.954 & \underline{0.922} & \underline{0.984} & \underline{0.953} & \textbf{0.962} & \textbf{0.729} & \underline{0.903} & \underline{0.856} & \textbf{0.892} & \underline{0.793} & \textbf{0.894} & \underline{2.0} \\
\bottomrule
\end{tabular}
}
\end{table*}

\subsection{Key Observations and Insights}
\paragraph{A. Superior Performance with Minimal Parameters.}

Tab.~\ref{tab:classification_auc} presents results across 10 diverse tasks. We observe that AnyMC3D (MII) and AnyMC3D (DINOv3) both reach 0.894 average AUC with ranks of 1.7 and 2.0, substantially outperforming all 3D classification baselines. Notably, this is achieved with only 1.3M trainable parameters per task: 10-40$\times$ fewer than competitive methods like MST (23.05M) and 40-50$\times$ fewer than 3D medical FMs like VoCo (50.49M). AnyMC3D secures first place on 7/10 tasks with MII and 4/10 tasks with DINOv3, demonstrating that properly adapted 2D FMs can achieve superior 3D classification results with unprecedented parameter efficiency.

\paragraph{B. General-Purpose FM Match Medical FM if Properly Adapted.}
In Tab.~\ref{tab:classification_auc}, AnyMC3D (DINOv3) achieves an average AUC of 0.894, matching the best medical FM, AnyMC3D (MII), and substantially surpassing other medical FMs. We also observe the same trend on the validation performance during training. As shown in Fig.~\ref{fig:pdac_val}a, DINOv3 underperforms MII by a large margin when both are used as frozen feature extractors. However, with our adaptation strategy, both DINOv3 and MII dramatically improve and achieve comparable performance after converging. These results demonstrate that while DINOv3 as a frozen feature extractor underperforms due to domain gap, its generic features possess remarkable transferability.

\textbf{Discussion: Rethinking Medical-Specific Pretraining.}
Our results indicate that generic visual features learned from natural images can be adapted to medical tasks as effectively as medical-specific features. This may be attributed to natural images having sharper boundaries and richer textures, which lead to more adaptable features that mitigate domain mismatch. Our findings also suggest that \textit{existing general-purpose 2D FMs, if properly adapted, may already be effective for 3D medical tasks, potentially reducing the need for expensive medical pretraining}.

\paragraph{C. General Yet Powerful: Surpassing Task-Optimized Baselines.}\label{42B}
Without task-specific tuning, we demonstrate that properly adapted 2D FMs (even non-medical FMs such as the DINO family) can outperform task-optimized baselines and win the 1st place in the VLM3D challenge.

\textbf{PDAC Early Detection (T5).} We compare against PanDx~\cite{liu2025pandx}, which ranked 1st in the PANORAMA challenge~\cite{alves2024panorama} by training with fine-grained segmentation labels. AnyMC3D (DINOv3), using only classification labels, outperforms PanDx by improving AUC from 0.949 to 0.962. By integrating auxiliary pixel-level supervision, performance further improves AUC from 0.962 to 0.973 (decoder architecture analysis in Appendix~\ref{app.a4}).

\begin{figure}[t]
    \centering
    \includegraphics[width=0.95\linewidth]{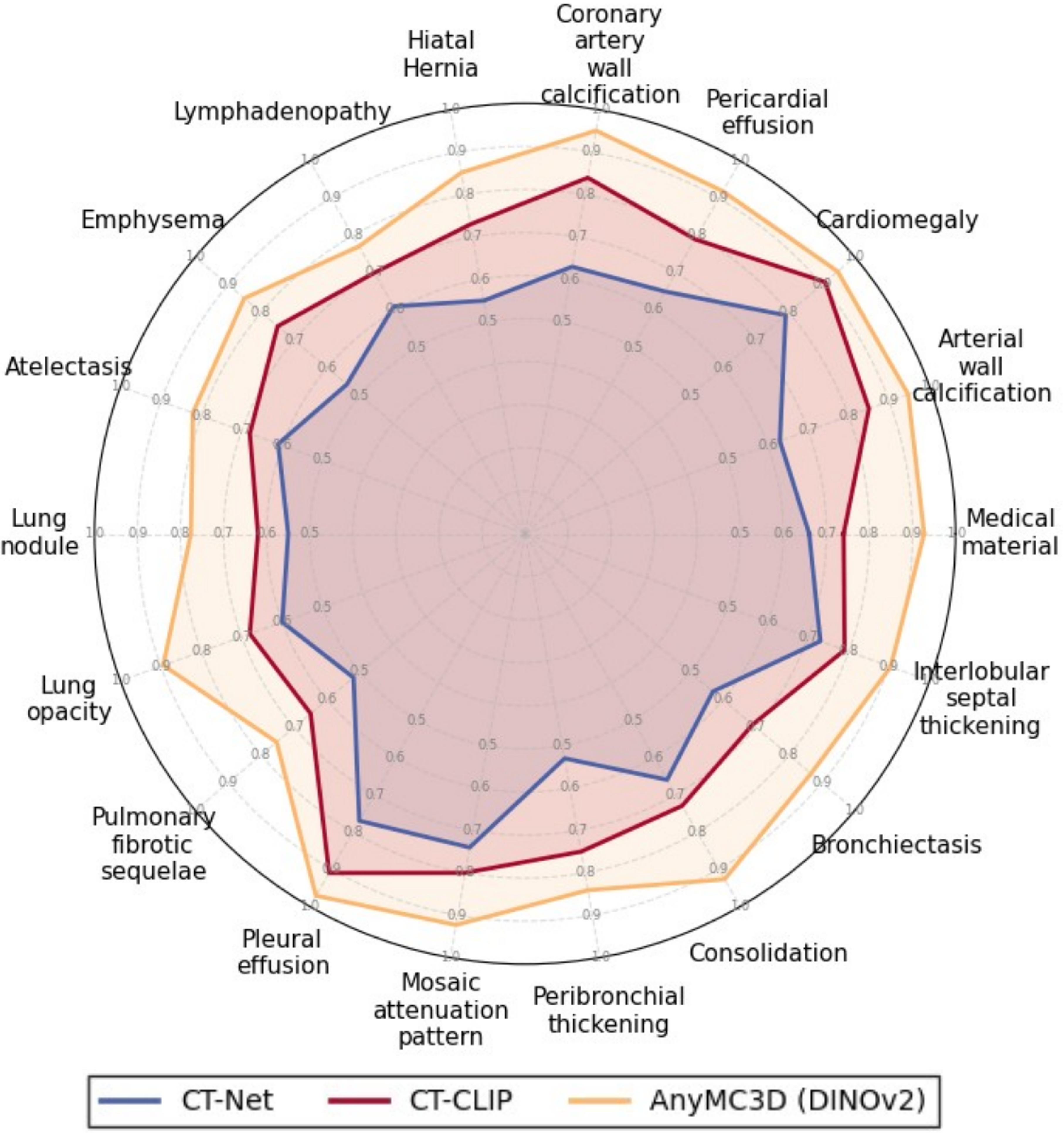}
    \caption{Classification performance of 18 chest CT abnormalities on CT-RATE dataset. Without medical-specific pretraining, AnyMC3D (DINOv2) outperforms (1) supervised baseline CT-Net and (2) vision-language FM, CT-CLIP, across all classes.}
    \label{fig:ct_rate}
\end{figure}

\textbf{Chest CT Multi-Abnormality (T11).} We evaluate on chest CT multi-abnormality classification using the \textbf{CT-RATE} dataset~\cite{hamamci2026generalist} with the \textbf{official data split} (47,149 training and 3,039 validation volumes). We compare AnyMC3D (DINOv2) against CT-Net~\cite{draelos2021machine} (supervised baseline) and CT-CLIP~\cite{hamamci2026generalist} (vision-language FM for chest CT). As shown in Fig.~\ref{fig:ct_rate}, AnyMC3D (DINOv2) consistently outperforms both baselines across all 18 diseases, improving mean AUC from 0.631 (CT-Net) and 0.748 (CT-CLIP) to 0.884. Per-finding AUC values are provided in Appendix~\ref{app.d}. We further validated AnyMC3D by attending the VLM3D challenge. With only about 0.5M trainable parameters, AnyMC3D won \underline{\textbf{1st place}} among 118 participants, beating other competing methods that performed expensive large-scale CT pretraining. This is a strong demonstration of AnyMC3D's generalizability, as challenge solutions are typically well-engineered and task-optimized.

\begin{figure}[t]
    \centering
    \includegraphics[width=0.95\linewidth]{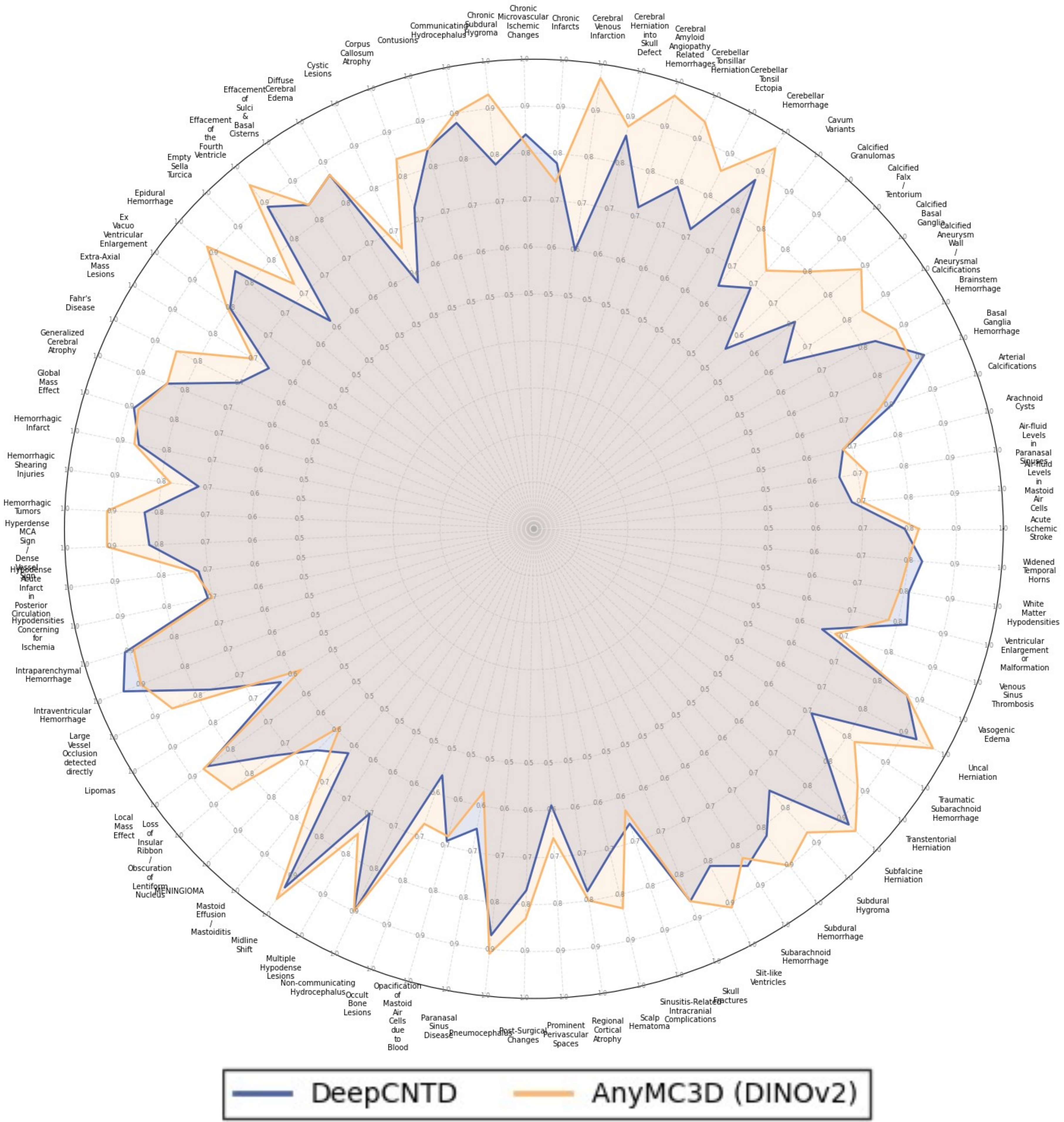}
    \caption{Performance comparison of 75 abnormalities for head CT emergency triage. We compare AnyMC3D (DINOv2) against the state-of-the-art neuroimaging FM, DeepCNTD.}
    \label{fig:headct}
\end{figure}

\textbf{Head CT Multi-Abnormality (T12).} We compare AnyMC3D (DINOv2) against the state-of-the-art head non-contrast CT FM for emergency triage, DeepCNTD~\cite{yoo2025non}, in classifying 75 head NCCT findings encompassing hemorrhagic, vascular, structural, traumatic, mass, and chronic abnormalities. As shown in Fig.~\ref{fig:headct}, our method improves the average AUC across these 75 findings from 0.768 to 0.820, demonstrating strong generalization capability. Notably, AnyMC3D achieves AUC $\ge$ 0.90 for 18 findings and 0.80 $\le$ AUC $<$ 0.90 for 33 findings, compared to 9 and 28 findings, respectively, for DeepCNTD.

\paragraph{D. Both Adaptation and Pretraining Matter.}\mbox{}\\
\textbf{(D1) Adaptation matters.} We assess the impact of adaptation by comparing the same medical FMs with different adaptation methods. Tab.~\ref{tab:classification_auc} shows that both MII and MedGemma can be substantially improved by replacing naive adaptation (linear probing) with AnyMC3D. Specifically, MII improves AUC from 0.785 (rank 8.7) to 0.894 (rank 1.7), and MedGemma improves from 0.690 (rank 10.9) to 0.866 (rank 5.2). Our results demonstrate that \textbf{(1)} adaptation strategy has a significant impact on downstream performance, and \textbf{(2)} \textit{linear probing}, the standard evaluation setup in most FM studies, is inadequate to unlock the full potential of FMs on medical tasks. 

\noindent\textbf{(D2) Pretraining matters.} We assess the impact of pretraining by comparing different FMs with the same adaptation strategy. Interestingly, AnyMC3D (DINOv3) outperforms AnyMC3D (MedGemma), though MedGemma was pretrained specifically for medical imaging. This demonstrates that medical pretraining alone does not guarantee superior results, and that general-domain FMs with effective adaptation can outperform domain-specific models.

\textbf{Discussion: Beyond Linear Probing.} Both 2D and 3D \textit{frozen} medical FMs yield suboptimal performance in our experiments. We believe this is because medical diagnostic findings are often extremely subtle, requiring task-specific adaptation of intermediate features rather than relying on frozen generic representations. Despite its importance, adaptation has been overlooked in previous FM research. Our study aims to raise the awareness of the community that \textit{neither high-quality pretraining nor effective adaptation alone is sufficient; both are essential to achieve state-of-the-art performance on medical imaging tasks}.

\begin{figure}[t]
    \centering
    \includegraphics[width=1\linewidth]{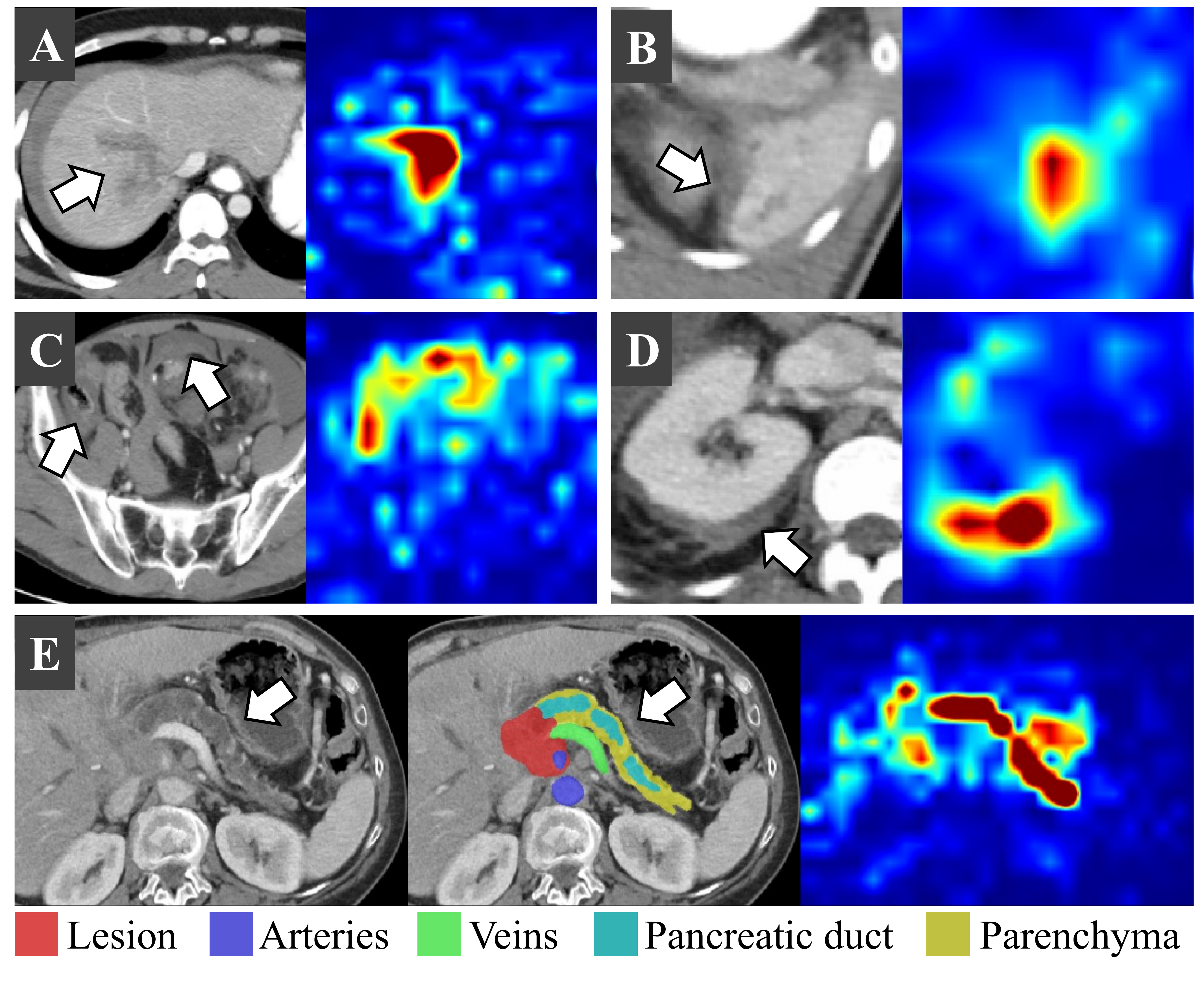}
    \caption{Examples of interpretable heatmaps. A: Liver injury, B: Spleen injury, C: Bowel injury, D: Kidney injury. E: The white arrow points to the critical secondary sign for PDAC.}
    \label{fig:heatmaps}
\end{figure}

\paragraph{E. Interpretable Heatmaps.}
Fig.~\ref{fig:heatmaps} shows the interpretable heatmaps generated by AnyMC3D. The heatmaps accurately localize diagnostically relevant regions, such as trauma injury sites across different organs (Fig.~\ref{fig:heatmaps}A-D). For PDAC early detection (Fig.~\ref{fig:heatmaps}E), our heatmap successfully highlights critical secondary signs, specifically pancreatic duct dilatation caused by downstream tumor obstruction. This is a key diagnostic indicator when the PDAC lesions cannot be clearly seen. These results demonstrate that adapted FMs learn clinically relevant features, providing interpretability for clinical deployment. We explore additional visualization methods, including Attention Rollout~\cite{abnar2020quantifying} and Gradient Attention Rollout, in the Appendix~\ref{app.g}.

\paragraph{F. Lightweight Adaptation Excels with Limited Data.}
In Fig.~\ref{fig:pdac_val}b, we compare AnyMC3D (DINOv3) against the 3D DenseNet across different training data regimes. AnyMC3D (DINOv3) dramatically outperforms DenseNet in low-data scenarios. With 20\% of the data (only 39 positive samples), AnyMC3D (DINOv3) achieves 0.924 AUC versus DenseNet's 0.741 (+0.18 AUC). Notably, AnyMC3D (DINOv3) with 20\% data surpasses DenseNet with 60\% data, demonstrating 3× data efficiency. This makes AnyMC3D ideal for scaling to new tasks, where abundant positive samples are typically not available in the early stage of data collection.

\begin{figure}[t]
    \centering
    \includegraphics[width=1\linewidth]{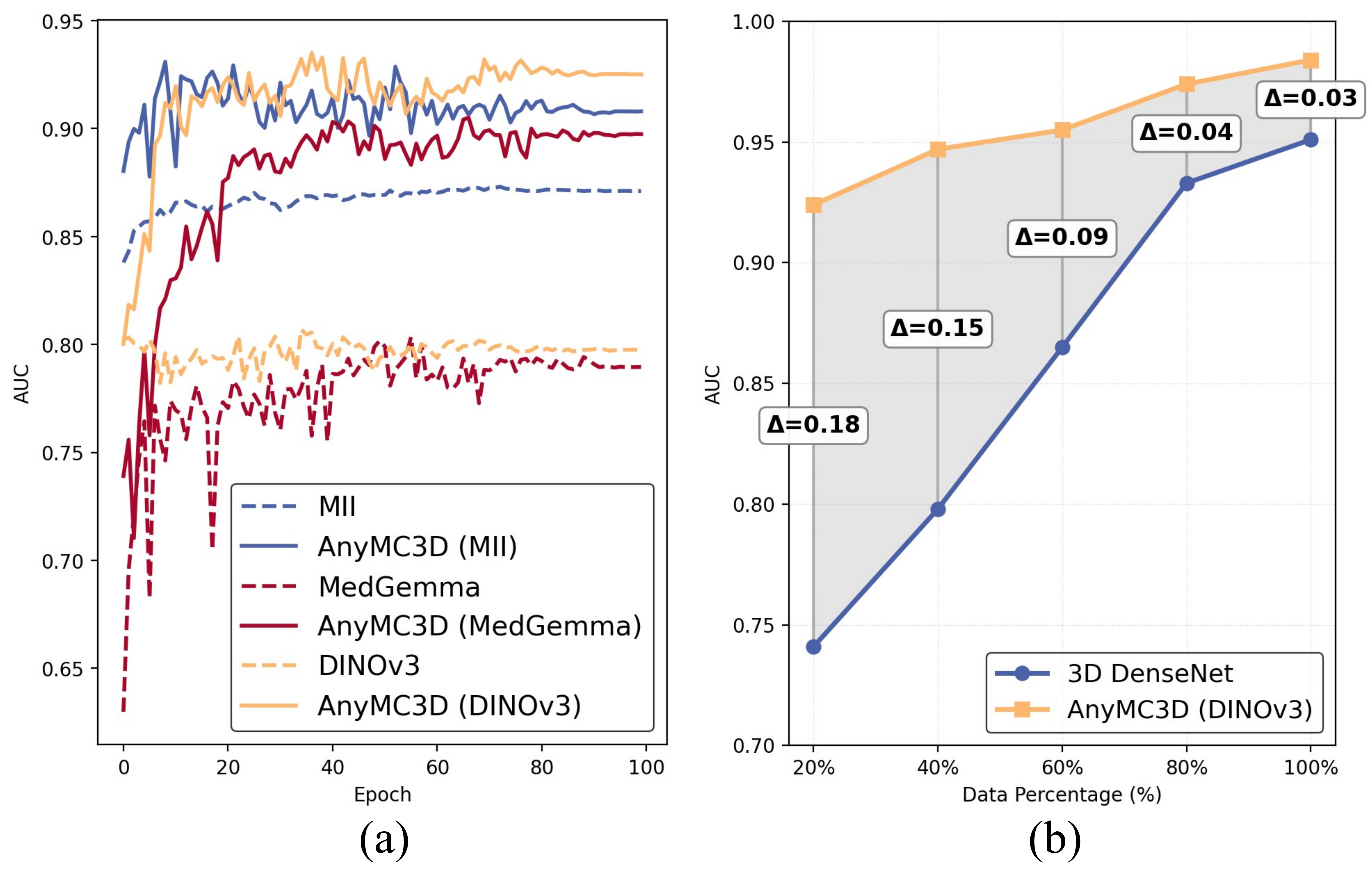}
    \caption{(a) Validation AUC on T5 by different adaptation methods. (b) Performance on T3 across different data regimes.}
    \label{fig:pdac_val}
\end{figure}

\paragraph{G. 2D Methods Surpass 3D Architectures.}
2D approaches consistently outperform 3D architectures across all parameter regimes. 3D models from scratch achieve only 0.683-0.833 average AUC with 11-33M parameters, with the best, DenseNet (0.833, 11.24M), trailing top 2D methods by 6.1 points. Pretrained 2D methods with slice fusion show clear advantages: MST achieves 0.869 (23.05M), substantially surpassing all 3D approaches. Even 3D medical FMs like MedicalNet (0.783, 46.16M) and VoCo (0.793, 50.49M) underperform despite larger parameter counts. This validates that leveraging pretrained 2D features with appropriate fusion outperforms 3D architectures for volumetric medical classification.

\textbf{Discussion: Why 2D Methods Excel for 3D Classification.}
This finding aligns with the empirical evidence that over the past five years, all the winning solutions for 3D classification challenges adopted the 2D/2.5D methods rather than 3D (see detailed summary in Appendix~\ref{app.e}). By contrast, 3D models, primarily nnU-Net variants~\cite{isensee2021nnu}, dominate 3D segmentation challenges~\cite{bassi2024touchstone,luo2025segrap2023,dorent2023crossmoda}. We hypothesize this discrepancy relates to fundamental task properties. Image-level classification can achieve robust predictions by aggregating slice-wise decisions, mirroring radiologists' workflow of scrolling through volumes slice-by-slice. In contrast, pixel-level segmentation demands fine-grained inter-slice relationships to accurately delineate 3D boundaries. Therefore, architectural choice may need to align with the spatial reasoning requirements of the target task rather than simply matching input dimensionality.

\paragraph{Ablation Studies.} We validate key design choices through three ablations (detailed in Appendix~\ref{app.f}).

\textbf{Slice Fusion Strategies.} We compare query-based attention pooling against average pooling, median pooling, and sequential modeling (LSTM~\cite{rsna2023kaggle}, Transformer~\cite{muller2025medical}). Tab.~\ref{tab:supp2} shows that attention pooling outperforms alternatives with minimal parameters, whereas sequential methods add computational overhead without gains.

\textbf{Backbone Size.} We compare different backbone sizes including ViT-S, ViT-B, and ViT-L. Tab.~\ref{tab:supp3} shows that larger backbones generally achieve better results. For new tasks, we recommend starting with ViT-S for efficient training and deployment, then scaling to larger models if needed.

\textbf{DINO Versions.} Tab.~\ref{tab:supp4} shows that DINOv2 and DINOv3 have negligible differences, indicating that pretraining improvements in DINOv3 do not translate to better 3D medical classification after adaptation.
\section{Conclusion}
\label{sec:discussion}
In this paper, we revisit scalable 3D medical classification and identify three critical pitfalls in previous research: data-regime bias, suboptimal adaptation, and insufficient task coverage. To address these pitfalls, we conduct comprehensive benchmarking and introduce AnyMC3D, a scalable 3D classifier adapted from 2D FMs that achieves state-of-the-art performance across a wide spectrum of tasks. Our work demonstrates the importance of both pretraining quality and adaptation strategy for FM performance, and highlights the strong potential of general-purpose 2D FMs for 3D medical imaging. We hope that AnyMC3D serves as a strong baseline for future 3D medical classification research.

\noindent\textbf{Limitations and Future Work.}
While our study reveals several key insights, some challenges remain. \textbf{First}, our observations are based on the FMs evaluated in this work; additional FMs may exhibit different adaptation characteristics. \textbf{Second}, all benchmark tasks use in-domain CT/MR data; FM generalization to out-of-domain modalities, \eg, PET scans, remains unexplored. \textbf{Third}, our auxiliary segmentation supervision relies on expensive pixel-level labels; weaker supervision such as bounding boxes or radiology reports could be explored via vision--language alignment. \textbf{Fourth}, DINOv3~\cite{simeoni2025dinov3} introduces Gram anchoring to enhance feature map quality. Although DINOv3 does not outperform DINOv2 for 3D medical classification (Tab.~\ref{tab:supp4}), its enhanced spatial features may benefit scalable 3D dense prediction tasks.

\paragraph{Disclaimer.} For research purposes only. Not for clinical use. This prototype is still under development and not yet commercially available. Future commercial availability cannot be guaranteed.

{
    \small
    \bibliographystyle{ieeenat_fullname}
    \bibliography{main}
}

\clearpage
\maketitlesupplementary
\appendix
\renewcommand{\thesection}{\Alph{section}}
\setcounter{section}{0}

\section*{Table of Contents}
\label{sec:toc}
\begin{enumerate}[label=\Alph*.]
    \item Implementation Details of AnyMC3D
    \item Dataset and Preprocessing
    \item Baseline Methods
    \item 1st Place in VLM3D Challenge
    \item Winners of 3D Classification Challenges
    \item Detailed Ablation Studies
    \item Attention Heatmaps
    \item Additional Evaluation Metrics
\end{enumerate}

\section{Implementation Details of AnyMC3D}\label{app.a}

\paragraph{A1. Model Configuration.}\label{app.a1}
AnyMC3D can be implemented with any transformer-based 2D foundation model (FM) backbone. We apply LoRA adapters to three components: (1) the patch embedding layer, (2) query, key, and value projection layers in self-attention, and (3) the output projection layer in self-attention. The LoRA rank $r$ is set to 8 and the scaling factor $\alpha$ is set to 16. The task query is initialized as a learnable parameter with values drawn from a truncated normal distribution with standard deviation 0.02. The classification head consists of a single linear layer. The final activation function is sigmoid for multi-label classification tasks and softmax for multi-class classification tasks.

\paragraph{A2. 2D Backbones for 3D Inputs.}\label{app.a2}
AnyMC3D processes 3D volumes through slice-wise encoding with 2D FMs (Alg.~\ref{alg:anymc3d_forward}). Each volume is partitioned into 2D slices along the highest-resolution axis, and the slice and batch dimensions are collapsed (reshaped to $(B \cdot S, C, H, W)$) for parallel processing. Single-channel medical slices are replicated three times to match the RGB input format and normalized with ImageNet statistics (mean=[0.485, 0.456, 0.406], std=[0.229, 0.224, 0.225]) to align with the FM's pretraining distribution. We also explored stacking three consecutive slices as a 2.5D input but found it performed comparably to single-slice replication.

\begin{algorithm}[t]
\caption{Forward Pass of AnyMC3D}
\label{alg:anymc3d_forward}
\DontPrintSemicolon
\SetKwInOut{Input}{Input}
\SetKwInOut{Output}{Output}
\SetKwInOut{Params}{Parameters}
\Input{3D volume $\mathbf{X} \in \mathbb{R}^{B \times C \times H \times W \times S}$ (or multi-view volumes), pretrained 2D FM $f_{\theta}$ (frozen), optional: return\_seg}
\Params{LoRA adapters $\{\psi^{(i)}\}$, task query $\mathbf{q}_{t}$, classification head $g_{\omega}$, optional view queries $\{\mathbf{q}^{(i)}\}$, optional 3D decoder $D$}
\Output{Classification logits $\mathbf{z} \in \mathbb{R}^{B \times K}$, optional segmentation logits $\mathbf{z}_{\text{seg}}$}
Initialize: $\text{views} \leftarrow [\,]$\\
\textcolor{gray}{\small\tcp{Process each view $i \in \{1, \ldots, V\}$}}
\For{each view $i$}{
    \textcolor{gray}{\small\tcp{Extract and prepare view}}
    $\mathbf{x}_i \leftarrow \text{ExtractView}_i(\mathbf{X})$\;
    $\hat{\mathbf{x}}_i \leftarrow \text{Rearrange}(\mathbf{x}_i, (B \cdot S, C, H, W))$\;
    $\hat{\mathbf{x}}_i \leftarrow \text{Normalize}(\hat{\mathbf{x}}_i)$\;
    
    \textcolor{gray}{\small\tcp{Slice-wise feature extraction}}
    $\mathbf{H}^{(i)} \leftarrow$ $\tilde{f}_{\theta,\psi^{(i)}}(\hat{\mathbf{x}}_i)$ (Eq.~\ref{slice_embed})\;
    
    \textcolor{gray}{\small\tcp{Query-based attention pooling}}
    $\mathbf{v}^{(i)} = \text{AttentionPool}(\mathbf{H}^{(i)}, \mathbf{q}^{(i)})$\;
    
    $\text{views} \leftarrow \text{views} \cup \{\mathbf{v}^{(i)}\}$\;
}
\textcolor{gray}{\small\tcp{Multi-view aggregation}}
\eIf{$V > 1$}{
    $\mathbf{V} = \text{Stack}([\mathbf{v}^{(1)}, \ldots, \mathbf{v}^{(V)}])$\;
    $\mathbf{v} = \text{AttentionPool}(\mathbf{V}, \mathbf{q}_{t})$\;
}{
    $\mathbf{v}=\mathbf{v}^{(1)} \quad$ \textcolor{gray}{\small\tcp{for single view, $\mathbf{q}^{(1)}=\mathbf{q}_{t}$}}
}
\textcolor{gray}{\small\tcp{Classification}}
$\mathbf{z} = g_{\omega}(\mathbf{v}) \in \mathbb{R}^{B \times K}$\;
\textcolor{gray}{\small\tcp{Optional: segmentation supervision}}
\If{return\_seg}{
    $\tilde{\mathbf{P}} \leftarrow$ Extract patch tokens from $\mathbf{H}^{(i)}$\;
    $\mathbf{P} \leftarrow \text{Reshape}(\tilde{\mathbf{P}}, (B, d, S, g_h, g_w))$ (Eq.~\ref{p3d})\;
    $\mathbf{z}_{\text{seg}} = D(\mathbf{P})$\;
}
\Return{$\mathbf{z}$, optional: $\mathbf{z}_{\text{seg}}$}
\end{algorithm}

\paragraph{A3. Training Details.}\label{app.a3}
During training, we employ focal loss $\mathcal{L}_{\text{focal}} = -\alpha (1 - p_t)^\gamma \log(p_t)$ to address class imbalance. We set $\gamma$ (the focusing parameter that down-weights easy examples) to 2 and $\alpha$ (the balancing parameter that addresses class imbalance) to 0.25. The batch size is 2 for all tasks except T10, where the batch size is 1 due to the large input dimension. We use a learning rate of 1e-4 with weight decay of 1e-5 for LoRA layers, and a learning rate of 1e-3 with weight decay of 1e-4 for the task query and classification head. The maximum number of training epochs is set to 100, and we select the best checkpoint based on validation AUC. During training, we apply strong data augmentation to both our method and all baselines. Our data augmentation strategy, adapted from nnU-Net~\cite{isensee2021nnu}, is applied directly to 3D images and includes random flipping along all three spatial axes (probability 0.5 each), random rotation ($\pm$30° per axis, probability 0.2), random zoom (0.7-1.4$\times$, probability 0.2), random affine translation ($\pm$10 voxels, probability 0.2), Gaussian noise (probability 0.25), Gaussian blur ($\sigma$=0.5-1.0, probability 0.2), brightness multiplication (0.75-1.25$\times$, probability 0.15), contrast augmentation (probability 0.15), low-resolution simulation (zoom 0.5-1.0$\times$, probability 0.2), and gamma correction ($\gamma$=0.7-1.5, probability 0.2-0.3 with/without image inversion).

\begin{figure}[t]
    \centering
    \includegraphics[width=1\linewidth]{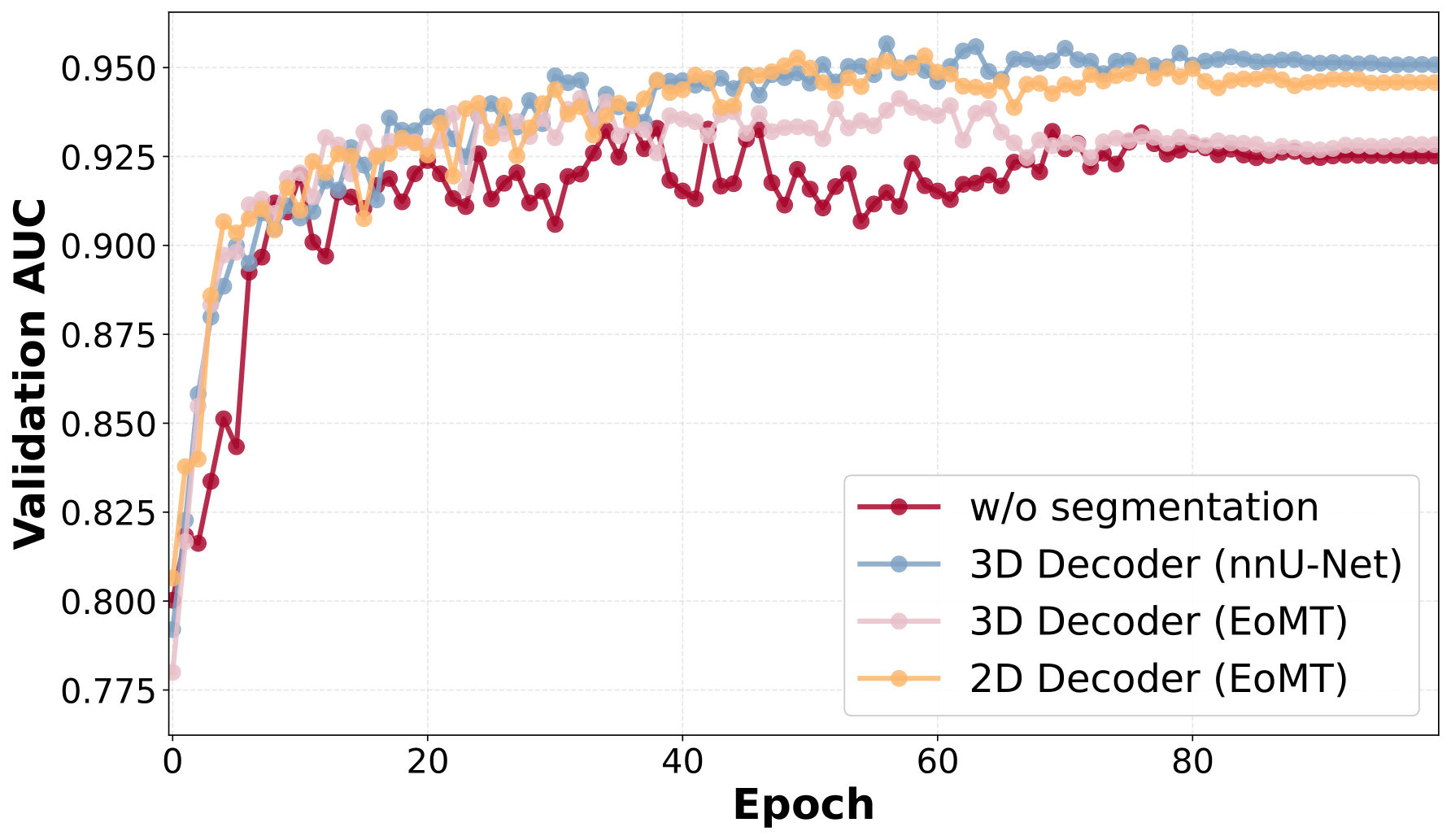}
            \caption{Validation AUC by different segmentation heads.}
    \label{supp1}
\end{figure}

\paragraph{A4. Choice of Segmentation Decoder Architecture.}\label{app.a4}
To incorporate pixel-level supervision, we employ a 3D decoder that upsamples the pseudo-3D token volume (Eq.~\ref{p3d}) into a 3D segmentation map. The decoder architecture follows the state-of-the-art 3D medical segmentation framework nnU-Net~\cite{isensee2021nnu}, consisting of consecutive blocks of 3D convolution, leaky ReLU activation (slope $p=0.01$), and 3D instance normalization. Following~\cite{isensee2021nnu}, we use the combination of Dice loss and cross-entropy loss for training. We also explored alternative decoder designs from EoMT~\cite{kerssies2025your}, a ViT-based segmentation model, in both 2D and 3D configurations. As shown in Fig.~\ref{supp1}, the nnU-Net-based decoder achieves the best performance, followed by the 2D EoMT decoder. These two designs also consistently outperform the baseline, demonstrating that regularizing patch tokens with segmentation effectively improves 3D classification.

\section{Dataset and Preprocessing}\label{app.b}
\paragraph{B1. Preprocessing.}\label{app.b1}
In Tab.~\ref{tab:dataset_supp}, we present the preprocessing steps, including normalization strategies, reshaped input sizes, data splits, and positive sample prevalence. All compared methods use identical preprocessing per task. For image normalization, we apply different strategies based on imaging modalities. For CT and CECT, we apply task-specific CT windows to highlight relevant pathologies and anatomies, then rescale to [0, 1]. For MRI, we apply z-score normalization to T8 and T10, and apply percentile clipping (0.5th to 99.5th percentiles) followed by rescaling to [0, 1] for T9. All images are resized into a consistent shape before feeding to the network. Following~\cite{isensee2021nnu}, the resized dimension is determined based on the median spacing of each dataset to minimize information loss from downsampling.

\begin{table*}[t]
\centering
\caption{Dataset summary across 12 tasks.}
\label{tab:dataset_supp}
\small
\setlength{\tabcolsep}{8pt}  
\begin{tabular}{ccccrrrrc}  
\toprule
Task & Modality & Normalization & Input size & \multicolumn{1}{c}{Total} & \multicolumn{1}{c}{Train} & \multicolumn{1}{c}{Val} & \multicolumn{1}{c}{Test} & \makecell{Pos.\\Ratio (\%)} \\
\midrule\midrule
\rowcolor{white} T1 & CECT & [-150, 250] & $288 \times 224 \times 126$ & 4,679 & 3,276 & 467 & 936 & 2.23 \\
\rowcolor{gray!15} T2 & CECT & [-150, 250] & $288 \times 256 \times 80$ & 4,701 & 3,290 & 471 & 940 & 10.09 \\
\rowcolor{white} T3 & CECT & [-150, 250] & $288 \times 128 \times 64$ & 4,677 & 3,274 & 469 & 934 & 5.99 \\
\rowcolor{gray!15} T4 & CECT & [-150, 250] & $160 \times 160 \times 70$ & 4,695 & 3,285 & 471 & 939 & 11.59 \\
\rowcolor{white} T5 & CECT & [-150, 250] & $432 \times 240 \times 70$ & 2,238 & 1,566 & 224 & 448 & 30.20 \\
\rowcolor{gray!15} T6 & CT & [-1000, 400] & $224 \times 224 \times 70$ & 1,140 & 684 & 115 & 341 & 76.32 \\
\rowcolor{white} T7 & CT & [-1000, 400] & $128 \times 128 \times 64$ & 5,668 & 3,604 & 1,032 & 1,032 & 14.04 \\
\rowcolor{gray!15} T8 & MRI & Z-score & $320 \times 320 \times 28$ & 12,159 & 11,143 & 508 & 508 & 38.14 \\
\rowcolor{white} T9 & MRI & PercentileClip & $320 \times 320 \times 28$ & 10,978 & 10,006 & 486 & 486 & 75.10 \\
\rowcolor{gray!15} T10 & MRI & Z-score & $320 \times 320 \times 28$ & 12,191 & 11,203 & 494 & 494 & 78.34 \\
\rowcolor{white} T11 & CT & A.S.L & $476 \times 476 \times 240$ & 50,188 & 45,149 & 2,000 & 3,039 & N/A \\
\rowcolor{gray!15} T12 & CT & B.T.B & $224 \times 192 \times 40$ & 29,476 & 23,657 & 2,930 & 2,889 & N/A \\
\bottomrule
\multicolumn{9}{l}{\small \textbf{A.S.L}: All, soft tissue, and lung windows ([-1000, 1000], [-150, 250], [-1000, 400]).} \\
\multicolumn{9}{l}{\small \textbf{B.T.B}: Bleeding, tissue, and bone windows.} \\
\end{tabular}
\end{table*}

\paragraph{B2. Dataset Description.}\label{app.b2}\mbox{}\\
\noindent\textbf{T1-T4. Abdominal Polytrauma.} 
We use the public RSNA Abdominal Trauma Detection (RATIC) dataset~\cite{rudie2024rsna}, which contains 4,703 contrast-enhanced CT (CECT) scans with annotations for four types of organ injuries: bowel (T1), liver (T2), kidney (T3), and spleen (T4). Two severity grades (Low: AAST~\cite{AAST_OIS_Abdominal_2025} grades 1-3; High: AAST grades 4-5) are annotated for liver, kidney, and spleen injuries, while only a single grade is provided for bowel injuries. We treat each organ injury as a binary classification task and train four separate models. For preprocessing, we crop the target organ region from each CECT scan using TotalSegmentator~\cite{wasserthal2023totalsegmentator}. We exclude cases where the limited field of view results in incomplete organ coverage, as these are unreliable for diagnostic decisions.

\noindent\textbf{T5. Pancreatic Cancer.} 
We use the public PANORAMA challenge dataset~\cite{alves2024panorama}, which contains 2,238 portal venous phase CECT scans from patients with pancreatic ductal adenocarcinoma (PDAC). We crop the pancreas region using pretrained segmentation models from the challenge baseline~\cite{alves2022fully}. This dataset includes both classification labels for PDAC and segmentation masks for six critical structures: PDAC lesion, veins, arteries, pancreatic duct, common bile duct, and pancreas parenchyma. In our major comparison (Tab.~\ref{tab:classification_auc}), we use only the classification labels to compare 3D classification methods. The segmentation labels are used for auxiliary supervision experiments reported in Sec.~\ref{42B} with the method presented in Sec.~\ref{aux}.

\noindent\textbf{T6. Lung Nodule Malignancy.} We collect a private dataset for classifying malignancy of biopsied high-risk lung nodules. The dataset includes 1,140 subjects from a single imaging site. The cohort is 47.55\% male and 52.45\% female, with a mean age of 67.30$\pm$12.11 years. Each subject has both diagnostic and biopsy CT scans. CT images are acquired on scanners from Siemens (72.25\%), GE (14.35\%), Philips (12.68\%), and Toshiba (0.72\%). A radiologist identifies and labels biopsied lung nodules on diagnostic CT images based on biopsy-needle positions in the corresponding biopsy CT images, yielding 1,140 nodules. Biopsy results yield binary labels: 76.32\% malignant and 23.68\% benign.

\noindent\textbf{T7. Lung Nodule Spiculation.} We collect a private dataset for classifying lung nodule spiculation, an important indicator of malignancy. It includes 3,884 CT scans from multiple imaging sites, acquired on scanners from GE (41.04\%), Siemens (34.76\%), Toshiba/Canon (12.74\%), Philips (2.70\%), and others (8.76\%). Three radiologists independently annotate spiculation for 6–30 mm solid nodules with ground truth defined by majority vote. In total, we obtain 5,668 nodules: 14.04\% spiculated and 85.96\% non-spiculated.

\noindent\textbf{T8. Bicep Tear.} 
We collect a large MR shoulder dataset comprising 11,828 subjects from two imaging sites. The cohort is 55.94\% male and 44.06\% female, with a mean age of 53.52$\pm$15.52 years. Each subject has both axial and sagittal MR scans. All scans are acquired with fat-saturation pulses. The magnetic field strength ranges from 0.7 T to 3.0 T, with a mean of 2.20$\pm$0.77 T. Most images (92.15\%) are acquired using Siemens scanners, followed by Philips (5.43\%), GE (1.72\%), and other manufacturers (0.70\%). The dataset includes biceps-tendon tear labels at three severity levels: no tear, tendinosis, and tear. Class distributions are 60.36\% no tear, 20.64\% tendinosis, and 19.01\% tear. For evaluation, we report the average of the three one-vs-rest AUC scores corresponding to the three classes.

\noindent\textbf{T9. Bursa Fluid.} 
We assemble an MR shoulder dataset of 10,126 subjects from two imaging sites. The cohort is 55.38\% male and 44.62\% female, with a mean age of 53.58$\pm$15.46 years. Each subject has axial, coronal, and sagittal fat-saturated MR scans. Field strengths range from 0.7 T to 3.0 T (mean 2.23$\pm$0.77 T). Scanners are predominantly Siemens (91.57\%), with Philips (5.20\%), GE (2.39\%), and others (0.75\%) comprising the remainder. Bursa-fluid labels are provided (no fluid vs. fluid present), with class proportions of 71.27\% and 28.73\%, respectively.

\noindent\textbf{T10. Labrum Tear.} 
Our MR shoulder dataset includes 11,816 subjects from two imaging sites. The population consists of 56.10\% males and 43.90\% females, with an average age of 53.37$\pm$15.52 years. Each subject undergoes coronal and sagittal fat-saturated MRI. The examinations are performed on scanners operating between 0.7 T and 3.0 T (mean 2.21$\pm$0.77 T). Siemens systems produce most scans (92.51\%), followed by Philips (5.07\%), GE (1.67\%), and other vendors (0.75\%). Labrum-tear status is annotated, with 67.15\% labeled as no tear and 32.85\% as tear.

\noindent\textbf{T11. Chest CT Multi-abnormality.} 
We use the public CT-RATE dataset~\cite{hamamci2026generalist} from Istanbul Medipol University, comprising 21,304 unique patients with 25,692 chest CT scans. The cohort ranges in age from 18 to 102 years, with a mean age of 48.8 years. The sex distribution is 41.6\% female and 58.4\% male. CT scans are acquired using three scanner manufacturers: Philips (61.5\%), Siemens (30.1\%), and PNMS (8.4\%). The number of slices per volume ranges from 100 to 600. Multi-abnormality labels for 18 distinct abnormalities are extracted from the corresponding radiology reports for each CT volume, including medical material, arterial wall calcification, cardiomegaly, pericardial effusion, coronary artery wall calcification, hiatal hernia, lymphadenopathy, emphysema, atelectasis, lung nodule, lung opacity, pulmonary fibrotic sequela, pleural effusion, mosaic attenuation pattern, peribronchial thickening, consolidation, bronchiectasis, and interlobular septal thickening.

\noindent\textbf{T12. Head CT Multi-finding.} We curate a large proprietary anonymized dataset of non-contrast head CT (NCCT) volumes for emergency triage, comprising 29,476 studies collected from nine centers across the U.S., Canada, China, and India, under ethics approvals with informed consent waived. Data are drawn from pre-established cohorts or retrospectively selected cases. NCCT scans are acquired using Siemens, GE, and Toshiba scanners. Exclusion criteria include patient age $<$18 years or absence of axial reconstruction. Seventy-five head NCCT findings, including hemorrhagic, vascular, structural, traumatic, mass, and chronic conditions, are extracted from radiology reports using large language models and subsequently verified by board-certified radiologists.

\section{Baseline Methods}\label{app.c}

\paragraph{C1. Implementation.}
This section describes implementation details for each baseline method. For methods with open-source repositories, we follow the original implementations and training hyperparameters. For others, we determine optimal hyperparameters through grid search.

\noindent\textbf{3D DenseNet.} 
We use the 3D DenseNet-121 implementation from MONAI\footnote{\scriptsize\url{https://github.com/Project-MONAI/MONAI}}.

\noindent\textbf{3D ResNet.} 
We use the 3D ResNet-18 implementation from MONAI, as it outperforms other variants (e.g., ResNet-50) in preliminary experiments.

\noindent\textbf{3D ConvNeXt.} 
We extend the 2D ConvNeXt~\cite{liu2022convnet} to 3D.

\noindent\textbf{M3T.} 
We follow the official implementation\footnote{\scriptsize\url{https://github.com/KVishnuVardhanR/M3T}}.

\noindent\textbf{MST.} 
We follow the official implementation\footnote{\scriptsize\url{https://github.com/mueller-franzes/MST}} and adopt the best-performing configuration from the paper: DINOv2-pretrained ViT-S as the backbone and a transformer without positional embeddings for slice fusion.

\noindent\textbf{RSNA-Kaggle.} 
We follow~\cite{rsna2023kaggle} and reimplement the model with 2D EfficientNet as the backbone and bidirectional LSTM for slice fusion. Following~\cite{rsna2022kaggle,rsna2023kaggle,rsna2024kaggle}, we stack consecutive slices as different channels to create 2.5D inputs. For fair comparison, we exclude model ensembling and remove the segmentation branch, as not all evaluated tasks include segmentation annotations.

\noindent\textbf{MedicalNet} 
is a 3D medical FM with a ResNet-50 backbone pretrained on large-scale 3D medical datasets~\cite{chen2019med3d}. We follow the official implementation\footnote{\scriptsize\url{https://github.com/Tencent/MedicalNet}} and evaluate three finetuning strategies: linear probing, LoRA adaptation, and full finetuning. In Tab.~\ref{tab:classification_auc}, we report full finetuning results as this achieves the best performance (Tab.~\ref{tab:supp1}).

\noindent\textbf{VoCo} 
is a 3D medical FM with a Swin-UNETR~\cite{hatamizadeh2021swin} backbone pretrained on large-scale 3D medical images~\cite{wu2024large}. We follow the official implementation\footnote{\scriptsize\url{https://github.com/Luffy03/Large-Scale-Medical}} using the VoComni\_B encoder for downstream classification. Similar to MedicalNet, we report full finetuning results as this outperforms other adaptation methods (Tab.~\ref{tab:supp1}).

\noindent\textbf{MedImageInsight (MII)} 
is a 2D FM pretrained on large-scale diverse medical images, including 2D modalities (fundus, pathology) and 2D slices from 3D imaging (CT, MRI)~\cite{codella2024medimageinsight}. We extract slice embeddings using the publicly available vision encoder\footnote{\scriptsize\url{https://huggingface.co/lion-ai/MedImageInsights}} as a frozen feature extractor, then aggregate them with our slice fusion method. We attempted full model finetuning but encountered severe overfitting due to the limited 3D training samples relative to the 360M-parameter DaViT backbone. We therefore use the frozen extraction setting, which is also the recommended configuration in the original paper.

\noindent\textbf{MedGemma} is a 2D medical FM built by finetuning the Gemma 3 vision encoder (SigLIP-400M) on over 33M medical image-text pairs, including 2D slices from CT and MRI~\cite{sellergren2025medgemma}. We use its vision encoder, MedSigLIP\footnote{\scriptsize\url{https://huggingface.co/google/medsiglip-448}}, as a frozen feature extractor to extract slice embeddings, which are then combined with our fusion method. The MII and MedGemma baselines provide valuable references for evaluating the out-of-the-box quality of 2D FM embeddings when adapted to 3D medical classification tasks through our slice fusion approach.

\paragraph{C2. Why Report Full Finetuning for 3D Medical FMs?}
In our main comparison (Tab.~\ref{tab:classification_auc}), we report full finetuning results for 3D medical FMs to represent their optimal performance. To justify this choice, we compare three adaptation strategies on T4 (Tab.~\ref{tab:supp1}): linear probing, LoRA, and full finetuning. For LoRA, we apply low-rank updates to convolutional layers in MedicalNet's ResNet-50 backbone and to query, key, and value projection layers in VoCo's Swin-UNETR backbone. Full finetuning achieves the best performance (MedicalNet: 0.899, VoCo: 0.919 AUC), significantly outperforming linear probing (0.654 and 0.702 AUC), while LoRA provides a parameter-efficient middle ground. We therefore report full finetuning results to represent 3D medical FMs at their optimal performance.

\begin{table}[ht]
\centering
\caption{Comparison of different finetuning strategies for 3D medical FMs on T4.}
\label{tab:supp1}
\small
\begin{tabular}{lc|cc}
\toprule
Method & Metric & MedicalNet & VoCo \\
\midrule\midrule
\multirow{2}{*}{LP} & Trainable Params (M) & 0.004 & 0.002 \\
& AUC & 0.654 & 0.702 \\
\midrule
\multirow{2}{*}{LoRA} & Trainable Params (M) & 1.14 & 0.07 \\
& AUC & 0.889 & 0.838 \\
\midrule
\multirow{2}{*}{Full} & Trainable Params (M) & 46.16 & 50.49 \\
& AUC & \textbf{0.899} & \textbf{0.919} \\
\bottomrule
\multicolumn{4}{l}{\footnotesize \textbf{LP}: Linear probing. \textbf{Full}: Full finetuning.} \\
\end{tabular}
\end{table}

\paragraph{C3. Task-Optimized Baselines.}
We compare against task-optimized baselines representing state-of-the-art performance for specific tasks, including challenge-winning solutions and specialized FMs tailored for particular clinical applications. This comparison rigorously tests whether \textit{AnyMC3D as a general framework can match or exceed specialized methods without task-specific designs}.

\noindent\textbf{PanDx (T5).} PanDx~\cite{liu2025pandx} ranked first in the PANORAMA challenge, achieving an AUROC of 0.9263 and AP of 0.7243. The method employs a two-stage coarse-to-fine approach: (1) a low-resolution segmentation model localizes the pancreatic region, and (2) a high-resolution model segments six PDAC-related structures and generates both patient-level likelihood scores and lesion-level detection maps. Both stages use nnU-Net~\cite{isensee2021nnu} trained on segmentation labels of pancreas-adjacent structures.

\noindent\textbf{CT-CLIP (T11)}~\cite{hamamci2026generalist} is a 3D FM trained via contrastive language-image pretraining that aligns CT volumes with report embeddings in a shared latent space. The model employs a 3D vision transformer as the image encoder and a text encoder to extract semantic features from radiology reports. During training, CT-CLIP maximizes cosine similarity between paired CT-report embeddings while minimizing similarity with negative pairs within each batch, enabling zero-shot abnormality detection. We compare against CT-CLIP's ClassFine variant, which finetunes a linear classifier on top of the pretrained frozen encoder, and the supervised baseline CT-Net~\cite{draelos2021machine}, a fully supervised 3D CNN trained directly on classification labels.

\noindent\textbf{DeepCNTD-Net (T12)}~\cite{yoo2025non} is a 3D neuroimaging FM that integrates two independently pretrained, task-specific vision networks through multi-modal fine-tuning with LLM-generated labels. The first network performs hemorrhage subtype segmentation using a 3D Dense U-Net optimized for five subtypes: intraparenchymal, subarachnoid, intraventricular, subdural, and epidural. The second network performs brain anatomy parcellation using a 3D U-Net with a multi-head design for segmenting left-hemisphere, supratentorial vs. infratentorial regions, and remaining brain structures. These pretrained networks are fused into a 3D DenseNet-based FM via feature-level integration, jointly encoding anatomical and pathological features.

\begin{figure}[ht]
    \centering
    \includegraphics[width=0.97\linewidth]{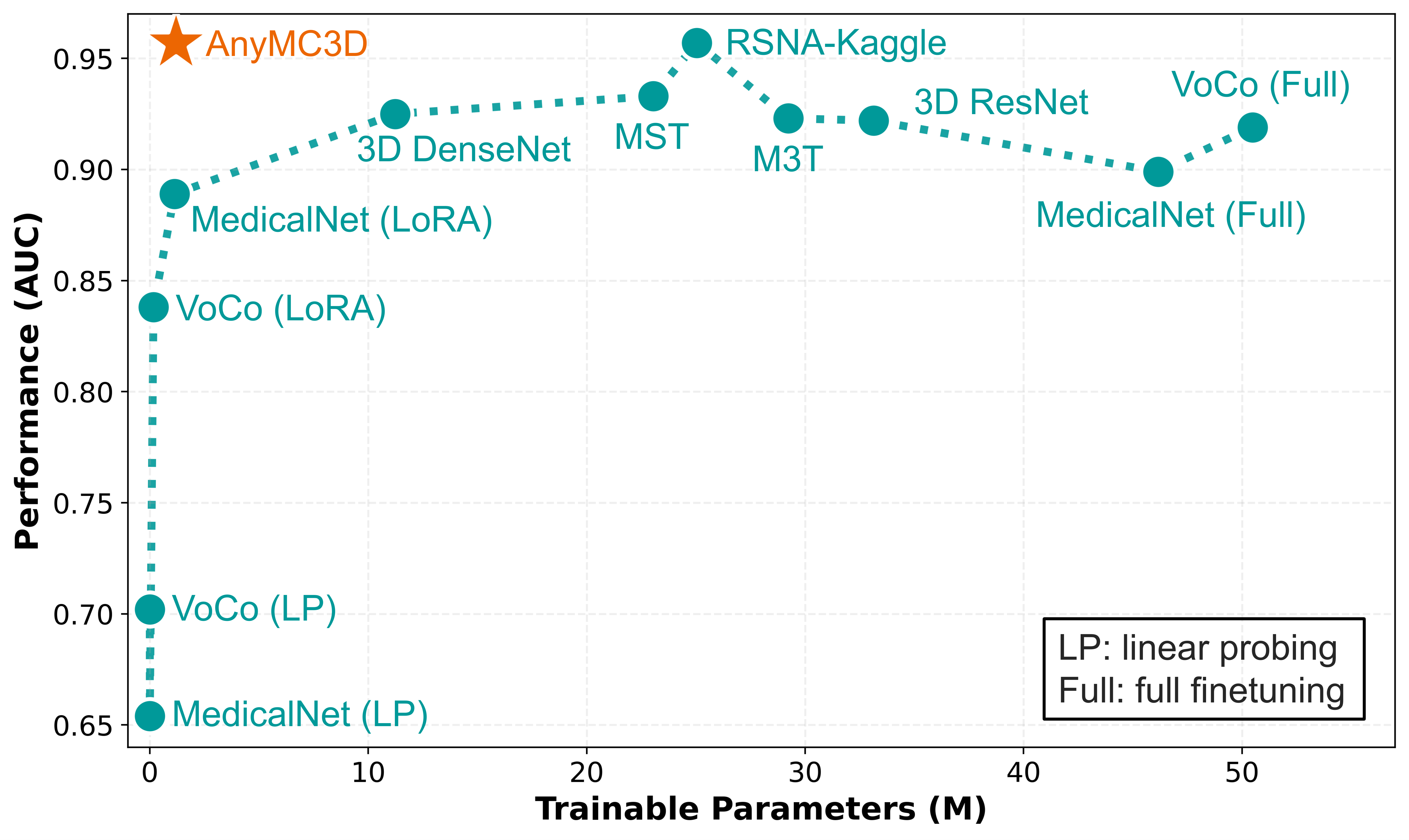}
            \caption{Performance and scalability on T4.}
    \label{param_eff}
\end{figure}

\paragraph{C4. Scalability vs. Performance.} 
Fig.~\ref{param_eff} illustrates the performance-scalability trade-off on T4. Existing approaches show clear compromises: 3D methods trained from scratch require 11--33M parameters for 0.92--0.93 AUC, while 2D/2.5D transfer learning methods need 23--29M parameters to reach 0.92--0.95 AUC. Fully finetuned 3D medical FMs (MedicalNet, VoCo) achieve 0.89--0.92 AUC with 46-50M parameters, but their parameter-efficient variants sacrifice significant performance (0.65--0.89 AUC) despite using $<$1M parameters. By contrast, AnyMC3D breaks this trade-off, achieving the highest performance (0.957 AUC) with only 1.32M trainable parameters.

\section{1st Place in VLM3D Challenge}\label{app.d}
To demonstrate AnyMC3D's out-of-the-box generalizability, we participated in the VLM3D challenge for multi-abnormality classification across 18 chest diseases on the CT-RATE dataset~\cite{hamamci2026generalist} (Appendix B2). Without bells and whistles, AnyMC3D achieved first place among 118 participants. For our submission, we use DINOv2 ViT-B as the 2D FM backbone. In this challenge, submissions are ranked based on three metrics: AUC, macro-F1 score, and clinically-weighted relevance gain (CRG) score. While AUC is threshold-agnostic, F1 and CRG require binary predictions at a fixed threshold of 0.5. Since focal loss training typically shifts the optimal operating point below 0.5, we apply model calibration to postprocess AnyMC3D outputs.

\noindent\textbf{Per-Finding AUC on CT-RATE.} Tab.~\ref{tab:ct_rate_auc} reports the per-finding AUC values for all 18 chest CT abnormalities.

\begin{table}[h]
\centering
\caption{Per-finding AUC of AnyMC3D (DINOv2) on the official validation split of CT-RATE.}
\label{tab:ct_rate_auc}
\small
\begin{tabular}{lc}
\toprule
Finding & AUC \\
\midrule
Medical material & 0.935 \\
Arterial wall calcification & 0.942 \\
Cardiomegaly & 0.948 \\
Pericardial effusion & 0.928 \\
Coronary artery wall calcification & 0.945 \\
Hiatal hernia & 0.863 \\
Lymphadenopathy & 0.779 \\
Emphysema & 0.851 \\
Atelectasis & 0.820 \\
Lung nodule & 0.789 \\
Lung opacity & 0.897 \\
Pulmonary fibrotic sequelae & 0.749 \\
Pleural effusion & 0.972 \\
Mosaic attenuation pattern & 0.922 \\
Peribronchial thickening & 0.845 \\
Consolidation & 0.929 \\
Bronchiectasis & 0.882 \\
Interlobular septal thickening & 0.909 \\
\midrule
Average & 0.884 \\
\bottomrule
\end{tabular}
\end{table}

\noindent\textbf{Platt Scaling Calibration.} We apply Platt scaling~\cite{platt1999probabilistic} to calibrate predictions per class:
\begin{align}
P_c^{'}=\sigma(z_c^{'}), \quad z_c^{'}=a_c z_c+b_c
\end{align}
where $P_c^{'}$ is the calibrated probability for class $c$, $z_c$ is the raw logit, $z_c^{'}$ is the calibrated logit, $\sigma$ is the sigmoid function, and $a_c, b_c$ are class-specific learned parameters. For each class, we first identify the raw operating point that maximizes F1 score on the validation set. We then fit a logistic regression model that takes the raw logit $z_c$ as input and the binary label as target, learning $a_c$ and $b_c$ to map the raw operating point to 0.5.

\begin{figure}[t]
    \centering
    \includegraphics[width=1\linewidth]{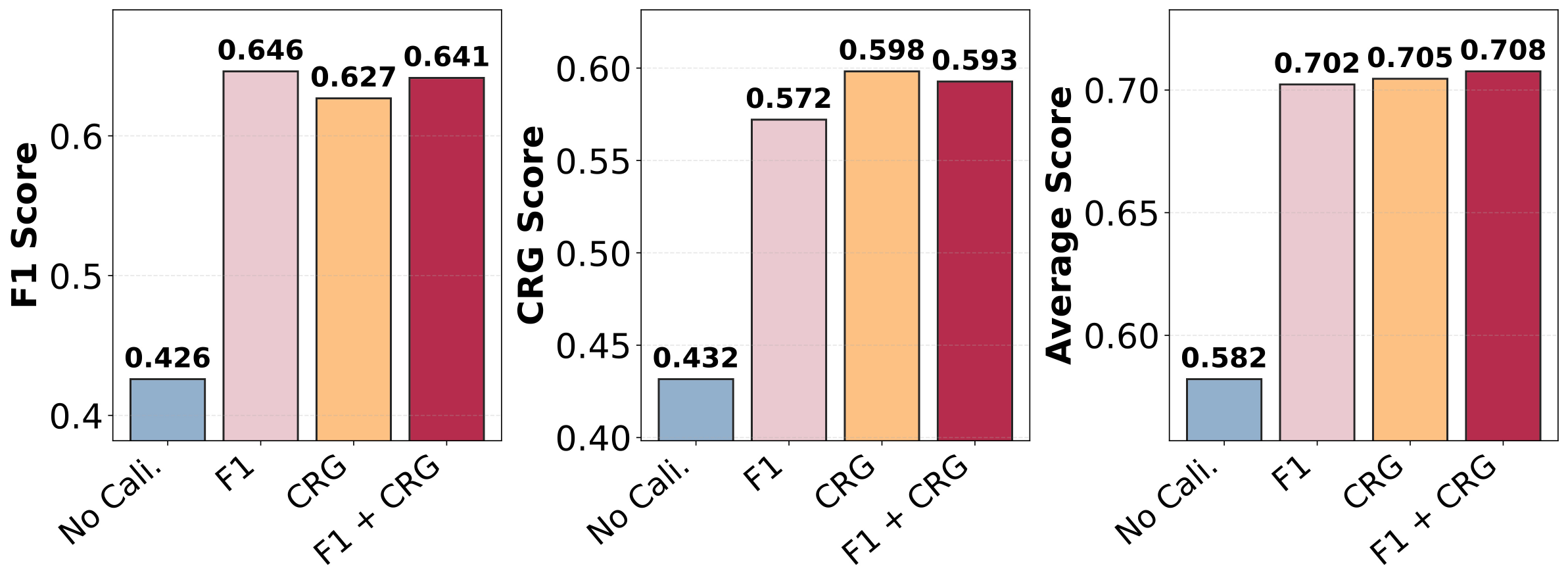}
    \caption{Evaluation metrics under different calibration strategies. Average denotes the mean of AUC, F1, and CRG scores.}
    \label{cali}
\end{figure}

\noindent\textbf{Calibration Strategy Analysis.} Fig.~\ref{cali} compares four calibration strategies: no calibration (No Cali.), optimizing for F1 only, CRG only, or balanced F1+CRG. All calibration strategies preserve AUC at 0.888 while substantially improving F1 and CRG scores. Since CRG weights true positives and false negatives by class prevalence, its optimal operating point differs from F1, which equally weights precision and recall. We optimize for F1 (pink) in our final submission to maximize F1 ranking. However, the balanced F1+CRG strategy may be preferable for real-world deployment where both metrics matter.

\noindent\textbf{Model Ensemble.} Our final submission ensembles 10 models by training 10 separate plugins (LoRA adapters, task-query, and classification head) with the shared DINOv2 backbone. Each model uses identical training configurations but different data splits that preserve per-class prevalence according to the global distribution. During inference, we efficiently switch plugins with the loaded FM backbone and average the calibrated logits across all 10 models.

\section{Winners of 3D Classification Challenges}\label{app.e}
We review top-performing solutions from recent 3D medical classification challenges. These winners emerge from highly competitive benchmarks and represent empirically validated strategies that outperformed strong baselines. We summarize key methodological takeaways from three consecutive RSNA-Kaggle challenges below.

\noindent\textbf{E1. RSNA-Kaggle 2022 (Cervical Spine Fracture).} The first-place solution~\cite{rsna2022kaggle} uses 2.5D CNN–RNN models for vertebra-level classification. For each vertebra, 15 slices are sampled along the $z$-axis and concatenated with neighboring slices and segmentation masks to form multi-channel 2D inputs. A 2D backbone (EfficientNet-V2-S~\cite{tan2021efficientnetv2} or ConvNeXt~\cite{liu2022convnet}) encodes each slice, and an LSTM head fuses features across slices for vertebra-level prediction. For patient-level prediction, the model jointly processes all seven vertebrae (7$\times$15 slices total) through the same 2.5D CNN+LSTM architecture, with final predictions obtained via ensemble across backbones and folds.

\noindent\textbf{E2. RSNA-Kaggle 2023 (Abdominal Trauma).}
The first-place solution~\cite{rsna2023kaggle} adopts a 2.5D slice-fusion approach. Each 96-slice volume is reorganized into 32 triplets of adjacent slices. A 2D CNN backbone (CoaT Lite~\cite{xu2021co} or EfficientNet-V2-S~\cite{tan2021efficientnetv2}) encodes each triplet, and a GRU sequence head models inter-slice dependencies. The model is trained with auxiliary segmentation heads and aggregates predictions via max pooling over slice logits. 

\noindent\textbf{E3. RSNA-Kaggle 2024 (Lumbar Spine Degenerative Classification).}
The first-place solution~\cite{rsna2024kaggle} employs a localize-then-classify pipeline. After 3D localization identifies level-wise coordinates, the classifier operates on multi-view crops (2.5D stacks of sagittal and axial slices). A 2D backbone (ConvNeXt-S~\cite{liu2022convnet} or EfficientNet-V2-S~\cite{tan2021efficientnetv2}) encodes each view, with a bidirectional LSTM and attention-based MIL fusing features across slices and views. Auxiliary attention losses regularize training, and ensemble predictions are obtained across backbones and folds.

\noindent\textbf{\underline{Takeaways and Motivation.}}
Winning solutions share common design patterns: (1) 2.5D representation via slice sampling or triplet formation, (2) 2D CNN backbones with explicit sequential modeling (LSTM/GRU or attention) for feature fusion, (3) auxiliary heads for training stabilization, and (4) multi-model ensembling for robust predictions. These findings demonstrate that 2D backbones with sequential fusion constitute the most effective approach for 3D medical classification in competitive settings. This motivates us to explore leveraging modern FMs' rich representations within the effective 2D+Fusion paradigm.

\section{Detailed Ablation Studies}\label{app.f}
\paragraph{F1. Impact of Slice Fusion Strategy.}\label{app.f1}
We evaluate different strategies for aggregating slice-level features into volume-level predictions on T5 (Tab.~\ref{tab:supp2}). Simple pooling operations (average, max, median) require no additional parameters but treat slices independently without modeling inter-slice relationships. Sequential methods like LSTM~\cite{rsna2024kaggle} and Transformer encoder~\cite{muller2025medical} explicitly model slice order but introduce substantial parameters (5.5M and 7.0M, respectively) and show mixed results, with LSTM underperforming (0.903 AUC) despite high parameter cost. Our query-based attention pooling achieves the best performance (0.962 AUC) with minimal parameters (0.001M), demonstrating that effective slice fusion does not require sequential modeling or a large parameter overhead. The learnable query automatically captures relevant cross-slice patterns through attention mechanisms, providing an optimal trade-off between performance and efficiency.


\begin{table}[h]
\centering
\caption{Comparison of slice fusion strategies on T5.}
\label{tab:supp2}
\small
\begin{tabular}{lccc}
\toprule
\multirow{2}{*}{Fusion Method} & \multirow{2}{*}{\makecell{Sequential \\ Modeling}} & \multirow{2}{*}{\makecell{Trainable \\ Params (M)}} & \multirow{2}{*}{AUC} \\
& & & \\
\midrule\midrule
Avg. pooling     & \Xmark & 0     & 0.958 \\
Max pooling      & \Xmark & 0     & 0.946 \\
Median pooling   & \Xmark & 0     & 0.944 \\
LSTM             & \Cmark & 5.5   & 0.903 \\
Transformer      & \Cmark & 7.0   & 0.950 \\
\midrule
Ours (Query-based) & \Xmark & \textbf{0.001} & \textbf{0.962} \\
\bottomrule
\end{tabular}
\end{table}

\paragraph{F2. Impact of Backbone Sizes.}\label{app.f2}
We evaluate three DINOv3 backbone sizes: ViT-S (21M), ViT-B (86M), and ViT-L (300M) across three representative tasks (Tab.~\ref{tab:supp3}). Results demonstrate that larger backbones consistently yield better performance across all tasks. ViT-L achieves the highest AUC on T3 (0.988), T5 (0.962), and T7 (0.903), outperforming ViT-S by margins of 0.008, 0.029, and 0.022, respectively. The performance gains come at the cost of increased trainable parameters: ViT-L requires 1.20M parameters compared to 0.23M for ViT-S. Notably, all backbone sizes maintain trainable parameters under 1.5M, demonstrating the parameter efficiency of our approach. For practical deployment, we recommend starting with ViT-S for rapid iteration and computational efficiency, then scaling to ViT-B or ViT-L when higher performance is required and computational resources permit.


\begin{table}[ht]
\centering
\caption{Comparison of DINOv3 backbone sizes across tasks.}
\label{tab:supp3}
\small
\begin{tabular}{l c ccc}
\toprule
\multirow{2}{*}[-0.5em]{Backbone} & \multirow{2}{*}[-0.5em]{\makecell{Trainable\\Params (M)}} & \multicolumn{3}{c}{AUC} \\
\cmidrule(lr){3-5}
&  & T3 & T5 & T7 \\
\midrule\midrule
ViT-S (21M)  & 0.23 & 0.980 & 0.933 & 0.881 \\
ViT-B (86M)  & 0.46 & 0.975 & 0.943 & 0.902 \\
ViT-L (300M) & 1.20 & \textbf{0.988} & \textbf{0.962} & \textbf{0.903} \\
\bottomrule
\end{tabular}
\end{table}

\paragraph{F3. Impact of DINO Versions.}\label{app.f3}
We compare DINOv2 and DINOv3 across four tasks (Tab.~\ref{tab:supp4}). Both versions achieve comparable performance with negligible differences, and neither consistently outperforms the other. For 3D classification tasks, the choice between DINO versions appears inconsequential. This may be explained by DINOv3's primary improvement over DINOv2: the introduction of a Gram anchoring mechanism~\cite{simeoni2025dinov3} that prevents the degradation of local (patch-level) features during long training periods. While this improvement does not translate to better performance for image-level tasks, i.e., classification, it may offer advantages for dense prediction tasks such as segmentation and detection that require fine-grained spatial features. Exploring DINOv3's potential for scalable 3D medical segmentation/detection remains a promising direction for future work, as discussed in our conclusion.

\begin{table}[ht]
\centering
\caption{Comparison of DINO versions on abdominal trauma classification tasks (T1-T4).}
\label{tab:supp4}
\small
\begin{tabular}{lcccc}
\toprule
\multirow{2}{*}[-0.5em]{Version} & \multicolumn{4}{c}{AUC} \\
\cmidrule(lr){2-5}
& T1 & T2 & T3 & T4 \\
\midrule\midrule
DINOv2 & 0.956 & 0.914 & 0.988 & 0.957 \\
DINOv3 & 0.954 & 0.922 & 0.984 & 0.953 \\
\bottomrule
\end{tabular}
\end{table}

\section{Attention Heatmaps}\label{app.g}
\begin{figure}[h]
    \centering
    \includegraphics[width=1\linewidth]{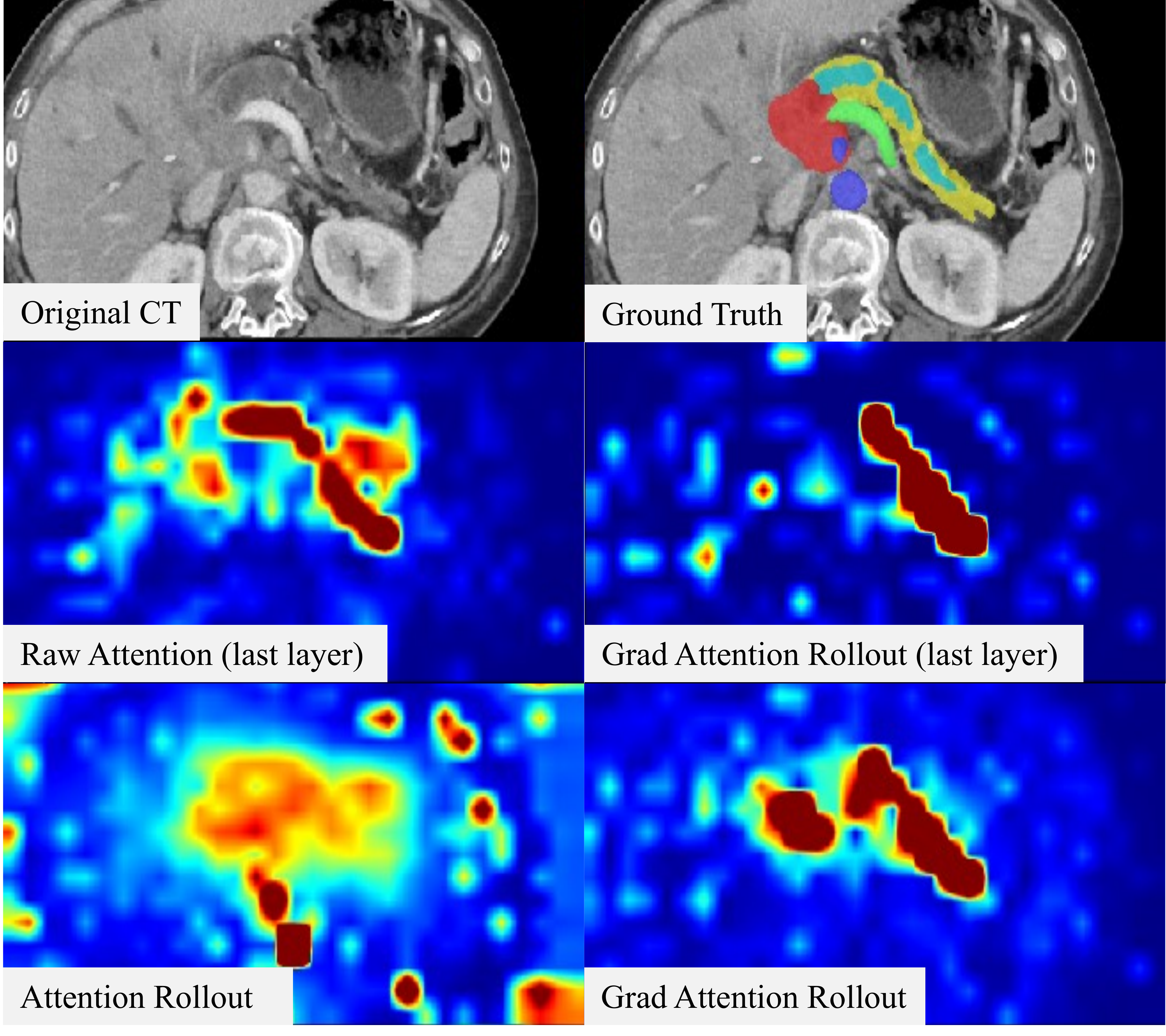}
            \caption{Heatmaps generated by different visualization methods.}
    \label{more_heatmaps}
\end{figure}

We explore multiple explainability methods to generate interpretable heatmaps with our ViT-based framework (Fig.~\ref{more_heatmaps}). Beyond raw attention maps from the last layer (shown in Sec.~\ref{sec:experiments}), we evaluate: (1) Attention Rollout~\cite{abnar2020quantifying}, which aggregates attention weights across all transformer layers to trace information flow; (2) Gradient Attention Rollout, which weights attention maps by gradients of the predicted class to highlight decision-relevant regions; and (3) Gradient Attention Rollout (last layer), which applies gradient weighting only to the final layer. 

Our comparison reveals three key observations. First, all methods successfully identify clinically relevant features, with high activation on pancreatic duct dilatation, a critical secondary sign for PDAC diagnosis. Second, Attention Rollout produces noisier heatmaps with diffuse activations, lacking class-specific guidance. Third, gradient-based methods generate more localized heatmaps by incorporating decision-level information, with the last-layer variant producing the most focused activations on discriminative anatomical structures. Overall, these methods offer complementary visualization options for AnyMC3D.

\section{Additional Evaluation Metrics}\label{app.h}

\begin{table*}[!t]
\centering
\caption{Subgroup analysis of trauma organ injury grading.}
\label{tab:organ_performance}
\setlength{\tabcolsep}{10pt}
\small
\begin{tabular}{llrcccc}
\toprule
\multirow{2}{*}{Task} & \multirow{2}{*}{Grading} & \multirow{2}{*}{\makecell{Sample Size\\\# Pos $|$ \# Neg}} & \multirow{2}{*}{AUROC} & \multirow{2}{*}{Accuracy} & \multirow{2}{*}{Sensitivity} & \multirow{2}{*}{Specificity} \\
& & & & & & \\
\midrule\midrule
Bowel (T1) & 0 vs. 1 & 21 $|$ 915 & 0.9543 & 0.8996 & 0.9524 & 0.8984 \\
\midrule
\multirow{3}{*}{Liver (T2)} 
& 0 vs. 1+2 & 94 $|$ 846 & 0.9219 & 0.8617 & 0.8511 & 0.8629 \\
& 0 vs. 1 & 76 $|$ 846 & 0.9049 & 0.8590 & 0.8158 & 0.8629 \\
& 0 vs. 2 & 18 $|$ 846 & 0.9934 & 0.8657 & 1.0000 & 0.8629 \\
\midrule
\multirow{3}{*}{Kidney (T3)}
& 0 vs. 1+2 & 55 $|$ 879 & 0.9842 & 0.9636 & 0.9636 & 0.9636 \\
& 0 vs. 1 & 34 $|$ 879 & 0.9775 & 0.9628 & 0.9412 & 0.9636 \\
& 0 vs. 2 & 21 $|$ 879 & 0.9949 & 0.9644 & 1.0000 & 0.9636 \\
\midrule
\multirow{3}{*}{Spleen (T4)}
& 0 vs. 1+2 & 109 $|$ 829 & 0.9527 & 0.9318 & 0.8807 & 0.9385 \\
& 0 vs. 1 & 63 $|$ 829 & 0.9281 & 0.9316 & 0.8413 & 0.9385 \\
& 0 vs. 2 & 46 $|$ 829 & 0.9864 & 0.9383 & 0.9348 & 0.9385 \\
\bottomrule
\end{tabular}
\end{table*}
\paragraph{H1. Choice of Evaluation Metrics.} 
We primarily use AUROC in Sec.~\ref{sec:experiments} because it evaluates ranking quality independent of classification thresholds, making it robust to class imbalance and enabling fair comparison across tasks with varying positive rates (2.23\% to 78.34\%). In contrast, accuracy, sensitivity, and specificity are threshold-dependent metrics that require operating point selection based on specific clinical priorities. For example, trauma screening may prioritize high sensitivity to avoid missing injuries, while diagnostic confirmation may require high specificity to reduce unnecessary interventions.

\paragraph{H2. Additional Metrics and Subgroup Analysis.}
Tab.~\ref{tab:organ_performance} provides a comprehensive evaluation beyond AUROC, including accuracy, sensitivity, and specificity across trauma grading scenarios using Youden's J statistic for threshold selection. While Sec.~\ref{sec:experiments} reports AUROC for binary classification (injury vs.\ no injury, i.e., 0 vs.\ 1+2), we present granular performance by additionally separating low-grade (0 vs.\ 1) and high-grade (0 vs.\ 2) scenarios.

Our method demonstrates exceptional performance for severe injuries (0 vs.\ 2), achieving AUROC values of 0.9934 (Liver), 0.9949 (Kidney), and 0.9864 (Spleen), with perfect sensitivity (1.000) for Liver and Kidney. This indicates the framework reliably identifies high-grade trauma without missing positive cases. For the binary classification task (0 vs.\ 1+2), performance remains strong with accuracy of 0.862-0.964 and well-balanced sensitivity-specificity trade-offs, where Kidney achieves perfectly balanced metrics (0.9636 for all three). 

Detection of low-grade injuries (0 vs.\ 1) proves more challenging, with sensitivity of 0.8158-0.9524, reflecting the inherent difficulty of identifying subtle trauma on CT imaging where findings may be ambiguous even to radiologists. Nevertheless, the method maintains high specificity (0.8629-0.9636) across all scenarios, correctly identifying non-injured cases and minimizing false alarms. These results confirm clinically relevant performance across the full spectrum of trauma severity with an appropriate sensitivity-specificity balance for different diagnostic scenarios.

\end{document}